\documentclass{article}

     \PassOptionsToPackage{numbers, compress}{natbib}

 \usepackage[final]{neurips_2020}

\usepackage[utf8]{inputenc} 
\usepackage[T1]{fontenc}    
\usepackage{url}            
\usepackage{booktabs}       
\usepackage{amsfonts}       
\usepackage{nicefrac}       
\usepackage{microtype}      
\usepackage{makecell}
\usepackage[pdftex]{graphicx}
\usepackage{color}
\usepackage{mathtools}
\usepackage[ruled]{algorithm}
\usepackage{algpseudocode}
\usepackage{subcaption}
\usepackage{upgreek}
\usepackage{wrapfig}
\usepackage{multirow}
\usepackage{amsmath}
\usepackage{enumitem}

\usepackage[colorlinks=true]{hyperref}
\hypersetup{citecolor=blue,linkcolor=red}

\newcommand{\minimize}{\mathop{\rm minimize}}

\newcommand{\argmin}{\mathop{\rm argmin}}

\newcommand{\bx}{\ensuremath{{\mathbf x}}}
\newcommand{\calX}{\ensuremath{{\mathcal X}}}
\newcommand{\calY}{\ensuremath{{\mathcal Y}}}
\newcommand{\calP}{\ensuremath{{\mathcal P}}}
\newcommand{\calD}{\ensuremath{{\mathcal D}}}
\newcommand{\calI}{\ensuremath{{\mathcal I}}}

\newtheorem{theorem}{Theorem}

\title{Robust Recursive Partitioning for Heterogeneous Treatment Effects with Uncertainty Quantification}

\author{%
	Hyun-Suk Lee\thanks{Equal contribution} \\
	Sejong University\\
	\texttt{hyunsuk@sejong.ac.kr} \\
	\And
	Yao Zhang$^*$ \\
	University of Cambridge \\
	\texttt{yz555@cam.ac.uk} \\
	\And
	William R. Zame \\
	UCLA \\
	\texttt{zame@econ.ucla.edu} \\
	\AND
	Cong Shen \\
	University of Virginia \\
	\texttt{cong@virginia.edu} \\
	\And
	Jang-Won Lee \\
	Yonsei University \\
	\texttt{jangwon@yonsei.ac.kr} \\
	\And
	Mihaela van der Schaar \\
	University of Cambridge \\
	UCLA \\
	The Alan Turing Institute \\
	\texttt{mv472@cam.ac.uk} \\  
}

\begin{document}
	
	\maketitle
	
	\begin{abstract}
		Subgroup analysis of treatment effects plays an important role in applications from medicine to public policy to recommender systems.  It allows physicians (for example) to identify groups of patients for whom a given  drug or treatment is likely to be effective and groups of patients for which it is not.  Most of the current methods of subgroup analysis begin with a particular algorithm for estimating individualized treatment effects (ITE) and identify subgroups by maximizing the differences across subgroups of the average treatment effect in each subgroup.  These approaches have several weaknesses: they rely on a particular algorithm for estimating ITE, they ignore (in)homogeneity within identified subgroups, and they do not produce good confidence estimates.  This paper develops a new method for subgroup analysis, R2P, that addresses all these weaknesses.  R2P uses an arbitrary, exogenously prescribed algorithm for estimating ITE  and quantifies the uncertainty of the ITE estimation, using a construction that is more robust than other methods.  Experiments using synthetic and semi-synthetic datasets (based on real data) demonstrate that R2P constructs partitions that are simultaneously more homogeneous within groups and more heterogeneous across groups than the partitions produced by other methods.  Moreover, because R2P can employ any ITE estimator, it also  produces much narrower confidence intervals with a prescribed coverage guarantee than other methods.  
	\end{abstract}
	
	\section{Introduction} \vspace{-16pt}
	The understanding of treatment effects plays  an important role -- especially in shaping interventions and treatments -- in areas from  clinical trials \cite{rothwell2005subgroup,zhou2017residual} to recommender systems \cite{lada2019observational} to  public policy \cite{grimmer2017estimating}.  In many settings, the relevant population is diverse, and different parts of the population display different reactions to treatment.  
	In such settings, {\em heterogeneous treatment effect (HTE) analysis}   -- also called  {\em subgroup analysis}--  is used to find subgroups consisting of subjects who have similar covariates and display similar treatment responses \cite{imai2013estimating,foster2011subgroup}.  The identification of subgroups is informative of itself; it also  improves the interpretation of treatment effects across the entire population and makes it possible to develop more effective interventions and treatments and to improve the design of further experiments.  In a clinical trial, for example, HTE analysis can identify subgroups of the population for which the studied treatment is effective, even when it is found to be ineffective for the population in general  \cite{hobbs2011hierarchical}.
	
	To identify subjects who have similar covariates and display similar treatment responses, it is necessary to create reliable estimates of the treatment responses of individual subjects; i.e. of {\em individualized treatment effects (ITE)}.  The state-of-the-art work on HTE proceeds by simultaneously estimating ITE and recursively partitioning the subject population    \cite{athey2016recursive,su2009subgroup,tran2019learning,johansson2018interpretable}.
	In these HTE methods, the criterion for partitioning is maximizing  the heterogeneity of  treatment effects {\em across} subgroups, using a sample mean estimator, under the assumption that treatment effects are homogeneous {\em within}  subgroups.  In particular, the population (or any previously identified subgroup) would be  partitioned into two subgroups provided that the sample means of these subgroups are sufficiently different, ignoring the possibility that  treatment effects  might be very heterogeneous within the groups identified.  Put differently, these methods focus on inter-group heterogeneity but ignore intra-group heterogeneity.\looseness=-1

	\setlength{\intextsep}{7pt}%
	\begin{wrapfigure}{r}{0.4\textwidth}
		\centering
		\includegraphics[width=1.\linewidth]{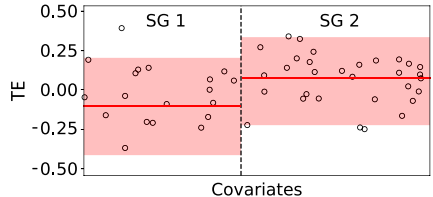}
		\caption{A toy example with two subgroups identified by HTE method in \cite{tran2019learning}. The solid red line shows the ITE estimation and their 95\% confidence interval is filled in red.}
		\label{fig:toy_example}
	\end{wrapfigure}
	
	An important problem with this approach is that, because it relies solely on inter-group heterogeneity based on sample means, it may lead to \textit{false discovery}.  To illustrate, consider the toy example depicted in Fig. \ref{fig:toy_example}.  
	In this example, the true treatment effect (shown on the vertical axis) was generated by iid random draws from the normal distribution having mean $0.0$ and standard deviation $0.1$.  In truth, the treatment under consideration is in fact {\em totally ineffective and innocuous};  on average, it has  {\em no effect at all} and the treatment effects are entirely uncorrelated with the single covariate (shown on the horizontal axis).  
	However if the observed data -- the realization of the random draws -- happens  to be the one shown in Fig. \ref{fig:toy_example}, standard methods will typically partition the population as shown in the figure, thereby ``discovering'' a segment of the population for whom the treatment is effective and a complementary segment where the treatment is dangerous.  Obviously, decisions based on such false discovery are useless -- or worse.
	Note that this false discovery occurs because, although the outcome variations between the two groups are indeed substantially different, the outcome variations within each group are just as different -- but the latter variation is entirely ignored in the creation of subgroups.\looseness=-1
	
	This paper proposes a robust recursive partitioning (R2P)\footnote{The code of R2P is available at: \url{https://bitbucket.org/mvdschaar/mlforhealthlabpub}.} method that avoids such false discovery.  R2P has several  distinctive characteristics. 
	\vspace{-\topsep}
	\begin{itemize}[leftmargin=*]\itemsep=0pt
		\item {\em R2P discovers interpretable subgroups in a way that is not tied to any {\em particular} ITE estimator.}  This is in sharp contrast with previous methods \cite{athey2016recursive,su2009subgroup,tran2019learning,johansson2018interpretable,seibold2016model}, each of which relies on a {\em specific} ITE estimator.  R2P can leverage {\em any} ITE estimator for subgroup analysis, e.g. an ITE estimator  based on Random Forest \cite{wager2018estimation}, or on multi-task Gaussian processes \cite{alaa2017bayesian} or on deep neural networks \cite{yoon2018ganite,shalit2017estimating,zhang2020learning}. This flexibility is important because no one ITE estimator is consistently the best in all different settings \cite{pmlr-v97-alaa19a}. 
			Furthermore, these ITE estimators are non-interpretable black-box models. R2P divides units into subgroups with respect to an interpretable tree-structure, and provides subgroup coverage guarantees for the ITE estimates in each subgroup. R2P enables these ITE estimators to produce trustworthy and interpretable ITE estimates in practice.
		
		
		\item  {\em R2P makes a conscious effort to guarantee homogeneity of treatment effects within each of the subgroups while maximizing heterogeneity across subgroups.} This is also different from previous methods, e.g., \cite{athey2016recursive,tran2019learning} where variation within the subgroup is largely ignored.
		
		\item {\em R2P produces  {\em confidence guarantees} with  narrower confidence intervals than previous methods.}  It accomplishes this by using methods of conformal prediction \cite{lei2018distribution} that produce valid confidence intervals.  Quantifying the uncertainty allows R2P to employ a novel  criterion we call {\em confident homogeneity} in order to create partitions that take account of both  heterogeneity across groups and homogeneity within groups. 
	\end{itemize}
	
		These characteristics make R2P both more reliable and more informative than the existing methods for subgroup analysis.
		Extensive experiments using synthetic and semi-synthetic datasets (based on real-world data) demonstrate that  R2P outperforms state-of-the-art methods by more robustly identifying subgroups while providing much narrower confidence intervals.

	
	\section{Robust Recursive Partitioning with Uncertainty Quantification}
	To highlight  the core design principles, we begin by introducing  robust recursive partitioning (R2P)  in the  regression setting;  we extend to the more complicated HTE setting in the next section.
	
	\subsection{Preliminaries}
	\vspace{-6pt}
	We consider a standard regression problem with a $d$-dimensional  covariate (input) space $\mathcal{X}\subseteq\mathbb{R}^d$ and a outcome space $\mathcal{Y}\subseteq\mathbb{R}$. We are given a dataset $\calD=\{(\bx_i,y_i)\}_{i=1}^n$, where, for the $i$-th sample, $\bx_i\in\calX$ is the vector of input covariates  and $y_i\in\calY$ is the outcome. We assume that  samples are independently drawn from an unknown distribution $\calP_{X,Y}$ on 
	$\calX\times\calY$. We are interested in estimating $\mu(\bx)=\mathbb{E}[Y|X=\bx]$, which is the mean outcome conditional on $\bx$. We denote the estimator by $\hat{\mu}:\calX\rightarrow\calY$;  $\hat{\mu}$ predicts an outcome $\hat{y}= \hat{\mu}(\bx)$ on the basis of the covariate information $\bx$. To quantify the uncertainty in the prediction, we apply the method of split conformal regression (SCR) \cite{lei2018distribution} to construct a confidence interval $\hat{C}$ that satisfies the rigorous frequentist guarantee in the finite-sample regime. (To the best of our knowledge, SCR is the simplest method that achieves this guarantee.)
	
	In SCR, we take as given a {\em miscoverage rate} $\alpha \in (0,1)$.  We split the samples in $\calD$ into a training set $\calI_1$ and a validation set $\calI_2$ that are disjoint and have the same size. We train the estimator 
	$\hat{\mu}^{\calI_1}$ on $\calI_1$ and  compute the residual of $\hat{\mu}^{\calI_1}$ on each sample in $\calI_2$. For a testing sample $\bx$ the confidence interval is given by
	\vspace{-6pt}
	\begin{equation}
		\label{eqn:conf_entire}
		\hat{C}^{\calI_1,\calI_2} (\bx) = \left[\hat{\mu}^{\textrm{lo}}(\bx),\hat{\mu}^{\textrm{up}}(\bx)\right]=\left[\hat{\mu}^{\calI_1}(\bx)-\hat{Q}_{1-\alpha}^{\calI_2},\hat{\mu}^{\calI_1}(\bx)+\hat{Q}_{1-\alpha}^{\calI_2}\right],
		\vspace{-6pt}
	\end{equation}
	where $\hat{Q}_{1-\alpha}^{\calI_2}$ is defined to be the $(1-\alpha)(1+1/|\calI_2|)$-th quantile of the set of residuals  $\{|y_i-\hat{\mu}^{\calI_1}(\bx_i)|\}_{i\in\calI_2}$. Assuming that the  training and testing samples are drawn exchangeably from $\calP_{X,Y}$,  the confidence intervals defined in (\ref{eqn:conf_entire})  satisfy the coverage guarantee $\mathbb{P}[y\in\hat{C}^{\calI_1,\calI_2}]\geq 1-\alpha$  \cite{lei2018distribution}.\footnote{Recall that assuming exchangeability is weaker than assuming iid.}  
	
	To illustrate, assume the miscoverage rate $\alpha$ is $0.05$ and we are given 1000 testing samples.  SCR prescribes a confidence interval for each sample in such a way that for at least 950 samples the prediction is within the prescribed confidence interval.  (We often say the sample is {\em covered}.)  However this coverage guarantee is marginal over the entire covariate space.  If we perform a  subgroup analysis that partitions the covariate space $\mathcal{X}$ into subgroups $\mathcal{X}_1, \mathcal{X}_2$ in such a way that $\mathcal{X}_1$ has 800  samples and $\mathcal{X}_2$ has 200 samples, it might be the case that 790 samples in $\mathcal{X}_1$ are covered but only $160$ samples in $\mathcal{X}_2$ are covered.  In this case,  80\% of the samples in $\mathcal{X}_2$ would be covered.  It seems obvious that such a situation is undesirable for subgroup analysis; we want to achieve the prescribed coverage rate for {\em each} subgroup, not just for the population as a whole.  R2P  overcomes this problem.   
	
	We begin by discussing how we use confidence intervals to quantify  outcome homogeneity within a subgroup. We then introduce our space partitioning algorithm and provide the required theoretical guarantee of subgroup coverage.
	
	\subsection{Partitioning for Robust Heterogeneity Analysis}
	\label{sec:criterion}
	Let $\Pi$ be a partition of the covariate space $\mathcal{X}$ into (disjoint) subgroups.  Write  $|\Pi|$ for the number of subgroups in the partition, $l_j$ for an element of $\Pi$ and $l(\bx;\Pi)$ for the subgroup that contains the sample $\bx$. Write $\calD_l=\{(\bx_i,y_i)\in\calD|\bx_i\in l\}$ for the samples whose covariates belong to the subgroup $l$. Note that when we restrict to covariates in the subgroup $l$, the samples are drawn from the truncated distribution $\calP_{X,Y}^l$ which is the distribution conditional on the requirement that the vector of covariates of samples lie in the subgroup $l$. 
	
	We evaluate  homogeneity within the subgroup $l$ by the concentration of outcome values for covariate vectors in the subgroup $l$. To do this, we apply SCR to the samples in  $\calD_l$ by splitting it into two sets, $\calI_1^l$ and $\calI_2^l$. Write $\hat{\mu}_l(\bx)$ denote the mean outcome model trained on  $\calI_1^l$. As in (\ref{eqn:conf_entire}), we  obtain the confidence interval $\hat{C}_l(\bx)$ for subgroup $l$ by setting the upper and lower endpoints to be
	$\hat{\mu}_l^{\textrm{up}}(\bx)=\hat{\mu}_l(\bx)+\hat{Q}_{1-\alpha}^{\calI_2^l}$ and $\hat{\mu}_l^{\textrm{lo}}(\bx)=\hat{\mu}_l(\bx)-\hat{Q}_{1-\alpha}^{\calI_2^l}$, respectively. (To avoid notational complications, omit reference to the subsets $\calI_1^l$ and $\calI_2^l$ hereafter; this should not cause confusion.  Throughout, we follow the convention that the confidence bound have been computed on the basis of the split.) To estimate the center of  the subgroup $l$, we use the average outcome $\hat{\mu}_{l,\textrm{mean}}=\mathbb{E}[\hat{\mu}_l(\bx)]$.  We define the {\em expected absolute deviation} in group $l$ to be
	$
	S_l=\mathbb{E}[v_l(\bx)],
	$
	where 
	\begin{equation}\label{equ:v_l}
		v_l(\bx)=\left(\hat{\mu}_{l,\textrm{mean}}-\hat{\mu}_l^{\textrm{up}}(\bx)\right)\mathbb{I}\left[\hat{\mu}_{l,\textrm{mean}}>\hat{\mu}_l^{\textrm{up}}(\bx)\right]
		+\left (\hat{\mu}_l^{\textrm{lo}}(\bx)-\hat{\mu}_{l,\textrm{mean}} \right)\mathbb{I}\left[\hat{\mu}_{l,\textrm{mean}}<\hat{\mu}_l^{\textrm{lo}}(\bx)\right].
	\end{equation}
	By definition,  $\hat{\mu}_l^{\textrm{up}}(\bx)$ is larger than 
	$\hat{\mu}_l^{\textrm{lo}}(\bx)$ provided the residual quantile $\hat{Q}_{1-\alpha}>0$. When the first indicator function is one, i.e. the average outcome (the group center) $\hat{\mu}_{l,\textrm{mean}}$ is larger than the upper bound $\hat{\mu}_l^{\textrm{up}}(\bx)$ at $\bx$, we are confident that the outcome value at $\bx$ is smaller than the group center (and perhaps smaller than the outcome value for many covariate vectors in $l$). Similarly, when the second indicator function is one, we are certain that the outcome value at $\bx$ is larger than the group center (and perhaps larger than the outcome value for many covariate vectors in $l$).  It is worth noting that when  $\hat{C}_l(\bx)=\big[\hat{\mu}_l^{\textrm{lo}}(\bx),\hat{\mu}_l^{\textrm{up}}(\bx)\big]$ contains the center $\hat{\mu}_{l,\textrm{mean}}$, both indicator functions are zero and $v_l(\bx)=0$. The quantity $v_l(\bx)$ evaluates the homogeneity in subgroup $l$ on the basis of the confidence interval for each 
	$\bx$ in $l$. (It is more conservative than the mean discrepancy $|\hat{\mu}_l(\bx)-\hat{\mu}_{l,\textrm{mean}}|$ for partitioning, and hence provides greater protection against false discovery because of  uncertainty.)
	
	\begin{figure}
		\centering
		\includegraphics[width=1.\textwidth]{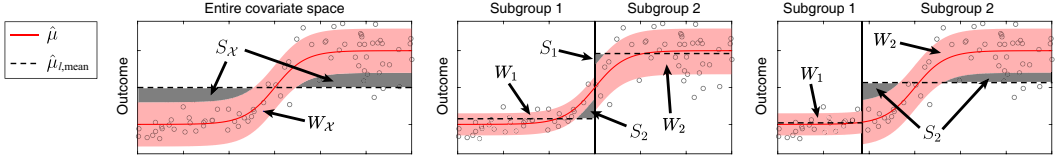}
		\caption{\textbf{Illustration of partitioning and impurity of the confident homogeneity.}  
			The regions shaded in red and gray (roughly) represent $W_l$ and $S_l$, respectively.
			Partitioning the heterogeneous covariate space (left panel) reduces its impurity of the confident homogeneity. The partition with the smaller impurity (middle panel) makes the heterogeneity across subgroups and the homogeneity within subgroups both stronger compared to others with the larger impurity (e.g., right panel). \label{fig:concept}}
		\vspace{-0.1in}
	\end{figure}
	
	However,  minimizing $S_l$ is not enough to maximize  subgroup homogeneity. If the intervals $\hat{C}_l(\bx)$ for all $\bx \in l$ are very wide and contain the average outcome $\hat{\mu}_{l,\textrm{mean}}$,  homogeneity can be very low even though $S_l=\mathbb{E}[v_l(\bx)]$ is zero. To resolve this issue, when partitioning the covariate space we jointly minimize $S_l$ and the expected confidence interval width $W_l=\mathbb{E}\big[|\hat{C}_l(\bx)|\big]$. We formalize the robust partitioning problem as
	\begin{equation}
		\label{eqn:objective}
		\minimize_{\Pi} \sum_{l\in\Pi} \lambda W_l + (1-\lambda) S_l,
	\end{equation}
	where $\lambda\in[0,1]$ is a hyperparameter that balances the impact of $W_l$ and $S_l$.  We call the weighted sum, $\lambda W_l + (1-\lambda) S_l$, the impurity of the \textit{confident homogeneity} for subgroup $l$. Fig. \ref{fig:concept} illustrates how minimizing the impurity of the confident homogeneity improves both homogeneity within subgroups and heterogeneity across subgroups. There may be more than one partition that achieves the minimum; because a larger number of subgroups is harder to interpret, we will choose a minimizer with the smallest number of subgroups.

	
	\subsection{Robust Recursive Partitioning Method with Confident Homogeneity}
	We can now describe our robust recursive partitioning (R2P) method for solving the optimization problem  \eqref{eqn:objective}.
	We begin with the trivial partition  $\Pi=\{\calX\}$.  
	We denote the set of subgroups whose objectives in \eqref{eqn:objective} can be potentially improved by $\Pi_c$.
	In the initialization step, we set $\Pi_c=\Pi$ and apply the SCR on $\calD$ to obtain the confidence intervals in (\ref{eqn:conf_entire}). Based on these intervals, we compute $\hat{W}_\calX$ and $\hat{S}_\calX$.
More generally, using the intervals for subgroup $l$, we can estimate $W_l$ and $S_l$ by
		\[
		\hat{W}_l=\frac{1}{N_2^l}\sum_{i\in\calI_2^l}|\hat{C}_l(\bx_i)| \ \  \text{and} \ \  \hat{S}_l=\frac{1}{N_2^l}\sum_{i\in\calI_2^l}v_l(\bx_i),
		\]
		respectively, where $N_2^l$ is the number of samples in the validation set $\calI_2^l$. 
	\begin{algorithm}[b]
		\caption{Robust Recursive Partitioning}\label{alg:algorithm}
		\begin{algorithmic}[1]
			\State \textbf{Input}: Samples $\calD=\{(\bx_i,y_i)\}_{i=1}^n$, miscoverage rate $\alpha\in(0,1)$, $\Pi=\{\calX\}$
			\State \textbf{Initialize}: $\Pi_{c}=\Pi$, split $\calD$ into $\calI_1^\calX$ and $\calI_2^\calX$, train $\hat{\mu}_\calX$, compute its confidence interval $\hat{C}_\calX$ using the split subsets, and obtain $\hat{W}_\calX$ and $\hat{S}_\calX$ using $\calI^\calX_2$
			\For{$l\in\Pi_c$}{}
			\State Obtain $\hat{W}^*_{l^\pm}$, $\hat{S}^*_{l^\pm}$, $i^*$, and $\phi^*$
			\If{$(1-\gamma)\left[\lambda\hat{W}_{l}+(1-\lambda)\hat{S}_l\right] \geq   \lambda\hat{W}^*_{l^\pm} + (1-\lambda)\hat{S}^*_{l^\pm}$}{}\Comment{\textit{Confident} criterion}
			\State Partition $l$ into $l^+(i^*,\phi^*)$ and $l^-(i^*,\phi^*)$
			\State $\Pi\leftarrow\Pi\setminus\{l\}$
			\State $\Pi\leftarrow\Pi\cup\{l^+(i^*,\phi^*),l^-(i^*,\phi^*)\}$ and $\Pi_c\leftarrow\Pi_c\cup\{l^+(i^*,\phi^*),l^-(i^*,\phi^*)\}$
			\EndIf
			\State $\Pi_c\leftarrow\Pi_c\setminus\{l\}$
			\EndFor
			\State \textbf{Output}: $\Pi$, $\hat{\mu}_\calX$, and $\hat{C}_l$ for all $l\in\Pi$
		\end{algorithmic}
	\end{algorithm}
	
	After initialization, we recursively partition the covariate space by splitting the subgroups in $\Pi_c$ with respect to the criterion in \eqref{eqn:objective}. To split each subgroup $l\in\Pi_c$, we first consider the two disjoint subsets from subgroup $l$ given by
	$l_k^+(\phi)=\{\bx\in l|x_k\geq\phi\} \textrm{ and } l_k^-(\phi)=\{\bx\in l|x_k<\phi\}$,
	where $\phi\in(x_k^{l,\textrm{min}},x_k^{l,\textrm{max}})$ is the threshold for splitting, $x_k$ is the $k$-th covariate element, and $x_k^{l,\textrm{min}}$ and $x_k^{l,\textrm{max}}$ are the minimum and maximum values of the $k$-th covariate within the subgroup $l$, respectively. We then apply SCR to each of these subsets. Specifically, we split the samples corresponding to $l_k^+(\phi)$ and $l_k^-(\phi)$ into  training and validation sets:
	$
	\calD_{l_k^+(\phi)} =\calI_1^{l_k^+(\phi)} \cup \calI_2^{l_k^+(\phi)} \textrm{ and }
	\calD_{l_k^-(\phi)}=\calI_1^{l_k^-(\phi)}\cup\calI_2^{l_k^-(\phi)},
	$
	where the split subsets are  $\calI_1^{l_k^+(\phi)}=\{(\bx_i,y_i)\in\calI_1^l:\bx_i\in {l_k^+(\phi)}\}$ and $\calI_2^{l_k^+(\phi)}=\{(\bx_i,y_i)\in\calI_2^l:\bx_i\in {l_k^+(\phi)}\}$ (same for ${l_k^-(\phi)}$).  To compute residuals, we do not train new estimators for $l_k^+(\phi)$ and $l_k^-(\phi)$; instead we use the previously trained estimator $\hat{\mu}_\calX$; this provides consistency of estimators across groups and within groups.  (It also avoids the enormous computational burden of  training new estimators for all the possible splits.) Using the residuals, we can construct the confidence intervals $\hat{C}_{l_k^+(\phi)}(\bx)$ and $\hat{C}_{l_k^-(\phi)}(\bx)$ and the associated quantities in the objective function, $\hat{W}_{l_k^+(\phi)},\hat{W}_{l_k^-(\phi)}, \hat{S}_{l_k^+(\phi)}$ and $\hat{S}_{l_k^-(\phi)}$. Then we find the optimal covariate $k^*_l$ and threshold $\phi^*_l$ for splitting subgroup $l$ as
	\[
	(k^*_l,\phi^*_l) = \argmin_{(k,\phi)} \lambda \left(\hat{W}_{l_k^+(\phi)}+\hat{W}_{l_k^-(\phi)} \right)+(1-\lambda)\left(\hat{S}_{l_k^+(\phi)}+\hat{S}_{l_k^-(\phi)} \right).
	\]
	For $(k^*_l,\phi^*_l)$, we compute $\hat{W}^*_{l^\pm}=\hat{W}_{l_{k^*}^+(\phi^*)}+\hat{W}_{l_{k^*}^-(\phi^*)}$ and $\hat{S}^*_{l^\pm}=\hat{S}_{l_{k^*}^+(\phi^*)}+\hat{S}_{l_{k^*}^-(\phi^*)}$. To improve the objective in \eqref{eqn:objective}, we split the subgroup $l$ into $l_{k^*}^+(\phi^*)$ and $l_{k^*}^-(\phi^*)$ only if the reduction in the impurity of the confident homogeneity is sufficiently large:
	\begin{equation}
		\label{eqn:confident_criterion}
		(1-\gamma)\left[\lambda\hat{W}_{l}+(1-\lambda)\hat{S}_l\right] \geq   \lambda\hat{W}^*_{l^\pm} + (1-\lambda)\hat{S}^*_{l^\pm}
	\end{equation}
	Here, $\gamma\in[0,1)$ is a hyperparameter for regularization. We refer to  \eqref{eqn:confident_criterion} as the \textit{confident criterion}.  With an appropriate choice of $\gamma$,  this criterion prevents overfitting, prevents the number of subgroups from becoming too large and prevents the size of each subgroup from becoming too small. Confident homogeneity does not necessarily improve as the group size shrinks because smaller groups lead to greater uncertainty.   This alleviates the issue of generalization to unseen data in HTE analysis \cite{athey2016recursive,tran2019learning}.   
	
	After the splitting decision, we remove $l$ from $\Pi_c$; if we have split $l$, we remove $l$ from $\Pi$ and add the two split sets to both $\Pi$ and $\Pi_c$.  We continue recursively until $\Pi_c$ is empty, at which point no further splitting is productive.  When the procedure stops, we will have obtained an estimator 
	$\mu_\calX$ and a partition $\Pi$ and for each $l \in \Pi$ we will have  corresponding confidence intervals $\hat{C}_l(\bx)=\left[\hat{\mu}_\calX(\bx)-\hat{Q}_{1-\alpha}^l,\hat{\mu}_\calX(\bx)+\hat{Q}_{1-\alpha}^l\right]$, where $\hat{Q}_{1-\alpha}^l$ is the $(1-\alpha)(1+1/|\calI_2^l|)$-th quantile of the set of the residuals on the validation set $\calI_2^l$ using $\hat{\mu}_\calX$, $\{|y_i-\hat{\mu}_\calX(\bx_i)|\}_{i\in\calI_2^l}$.  The following theorem guarantees that  the R2P partition Algorithm~\ref{alg:algorithm} provides a valid confidence interval  $\hat{C}_l$ for each subgroup  $l \in \Pi$; this is exactly what a user would want in subgroup analysis.  The proof is  provided  in the supplementary material.
	
	\begin{theorem}\label{thm:thm1}
		Given a prescribed miscoverage rate $\alpha$, the created partition $\Pi$,  estimator $\hat{\mu}_\calX$, and  confidence interval function $\hat{C}_l(\cdot)$ have the following property: for each $l \in \Pi$ and for new samples $(\bx,y)$ drawn from the truncated distribution $\calP_{X,Y}^l$, we have
		$\mathbb{P}[y\in\hat{C}_l({\bx})] \geq 1-\alpha$.
	\end{theorem}
	
	\vspace{-0.1in}
	\section{Robust Recursive Partitioning for Heterogeneous Treatment Effects} \vspace{-6pt}
	In this section, we extend the R2P method to the setting of HTE estimation, resulting in the \textit{R2P-HTE} design as detailed below.
	
	
	\begin{paragraph}{Heterogeneous Treatment Effect Model}
		We consider a setup with $n$ units (samples),
		For the unit $i\in\{1,2,...,n\}$, there exists a pair of potential outcomes, $Y_i(1)$ and $Y_i(0)$ that are independently drawn from an unknown distribution, where 0 and 1 represent whether the unit is treated or not, respectively. We define the treatment indicator as $t_i\in\{0,1\}$, where $t_i=1$ and $0$ indicate the unit $i$ is treated and untreated, respectively.
		The outcome for unit $i$ is realized as the potential outcome corresponding to its treatment indicator $y_i=Y_i(t_i)$. A dataset is given as $\calD_{\textrm{HTE}}=\{(\bx_i,t_i,y_i)\}_{i=1}^n$. The ITE for a given $\bx$ is defined as $\tau(\bx)=\mathbb{E}[Y(1)-Y(0)|X=\bx]$. Since the ITE is defined as the expected difference between the two potential outcomes $Y(1)$ and $Y(0)$, its estimator $\hat{\tau}(\bx)$ is given as the contrast between two regression models:  
		$\hat{\mu}^0(\bx)$ for the conditional non-treated outcome $\mathbb{E}[Y(0)|X=\bx]$, and $\hat{\mu}^1(\bx)$ for the conditional treated outcome $\mathbb{E}[Y(1)|X=\bx]$.
	\end{paragraph}
	
	\begin{paragraph}{R2P-HTE}
		We adapt R2P to HTE estimation by constructing the quantities $W_l$ and $S_l$ in \eqref{eqn:objective} based on the ITE estimator $\hat{\tau}(\bx)$. To construct an ITE estimator, many popular machine learning models have been considered in the literature \cite{athey2016recursive,wager2018estimation,alaa2017bayesian,yoon2018ganite,shalit2017estimating,zhang2020learning}. R2P-HTE can use one of these models to parameterize the outcome models $\hat{\mu}^0(\bx)$ and $\hat{\mu}^1(\bx)$. We set the target coverage rate of $\hat{\mu}^0(\bx)$ and $\hat{\mu}^1(\bx)$ as $\sqrt{1-\alpha}$.  As in the previous section, we can construct a confidence interval for each estimator by using the split conformal regression.
		Let us denote the $\sqrt{1-\alpha}$ confidence intervals for $Y(1)$ and $Y(0)$ by
		$
		\hat{C}^1(\bx)=\left[\hat{\mu}^1(\bx)-\hat{Q}^1_{\sqrt{1-\alpha}},\hat{\mu}^1(\bx)+\hat{Q}^1_{\sqrt{1-\alpha}}\right]
		$ and 
		$
		\hat{C}^0(\bx)=\left[\hat{\mu}^0(\bx)-\hat{Q}^0_{\sqrt{1-\alpha}},\hat{\mu}^0(\bx)+\hat{Q}^0_{\sqrt{1-\alpha}}\right],
		$
		respectively.
		We set the confidence interval for $\tau(\bx)$ to be
		\[
		\hat{C}^\tau(\bx)=\left[\hat{\mu}^1(\bx) - \hat{\mu}^0(\bx)-\hat{Q}^1_{\sqrt{1-\alpha}}-\hat{Q}^0_{\sqrt{1-\alpha}},\hat{\mu}^1(\bx) - \hat{\mu}^0(\bx)+\hat{Q}^1_{\sqrt{1-\alpha}}+\hat{Q}^0_{\sqrt{1-\alpha}}\right].
		\]
		This confidence interval ensures the coverage rate $1-\alpha$ for the estimated ITE $\hat{\tau}(\bx)=\hat{\mu}^1(\bx)-\hat{\mu}^0(\bx)$, because its upper endpoint is given as the difference between the upper endpoint of $\hat{C}^1$ and the lower endpoint of $\hat{C}^0$, and its lower endpoint is given as the difference between the lower endpoint of $\hat{C}^1$ and the upper endpoint of $\hat{C}^0$. If the coverage rates for $\hat{C}^1$ and $\hat{C}^0$ are $\sqrt{1-\alpha}$, the coverage rate for $\hat{C}^\tau$ will be $1-\alpha$.
		
		From the ITE estimator $\hat{\tau}(\bx)$ and its confidence interval $\hat{C}^\tau_l(\bx)$ for each subgroup $l$, we can calculate $W_l$ and $S_l$ and  adapt the R2P method to HTE estimation. The robust partitioning problem for HTE in \eqref{eqn:objective} is solved by applying the R2P method in Algorithm \ref{alg:algorithm}, with two minor changes: 1) each sample in the HTE dataset is a triple $(\bx_i,t_i,y_i)$ consisting of the covariate vector, the treatment indicator, and the observed outcome; 2) the outcome model $\hat{\mu}(\bx)$ in R2P is replaced by the ITE estimator 
		$\hat{\tau}(\bx)$.  As before, we show that this procedure achieves the specified coverage guarantee.   The proof is  provided in the supplementary material.
		
		\begin{theorem}\label{thm2}
			Given a prescribed miscoverage rate $\alpha$, the created partition $\Pi$,  estimator $\hat{\tau}_\calX$, and  confidence interval function $\hat{C}_l^\tau(\cdot)$ have the following property: for each $l \in \Pi$ and for new samples $(\bx,\tau)$ drawn from the truncated distribution of the subgroup $l$, we have  $\mathbb{P}[\tau\in\hat{C}_l^\tau({\bx})] \geq 1-\alpha$.
		\end{theorem}
	\end{paragraph}

\begin{paragraph}{Well-identified subgroups and false discovery} Theorem \ref{thm2} guarantees that confidence intervals achieve the required finite sample coverage for the ITE estimates in each subgroup, regardless of how (in)accurate the underlying ITE estimator $\hat{\tau}_\calX$ is. If the confidence intervals exhibit large overlap across the constructed subgroups, we can conclude that the constructed subgroups are not well-identified. Conversely, if the confidence intervals have little or no overlap across subgroups, we conclude that the subgroups are well-identified. Given the theoretical guarantee of the confidence intervals in R2P, the subgroups are robust against false discoveries if they are well-identified. \end{paragraph}
	

	\vspace{-0.1in}
	\section{Related Work} \vspace{-6pt}
	Subgroup analysis methods with recursive partitioning have been widely studied based on regression trees (RT) \cite{athey2016recursive,su2009subgroup,tran2019learning,johansson2018interpretable}.
	In these methods, the subgroups (i.e., leaves in the tree structure) are constructed and the individualized outcome or treatment effects are estimated by the corresponding sample mean estimator to the leaf for given covariates.
	To overcome the limitations of the traditional trees to represent the non-linearity such as interactions between treatment and covariates \cite{doove2014comparison},
	a parametric model is integrated into regression trees for subgroup analysis \cite{seibold2016model}.
	However, such approach can be used only for the limited types of estimator models, which is particularly undesirable since advanced causal inference models based on deep neural networks or multi-task Gaussian processes have been studied which outperform the traditional estimators \cite{alaa2017bayesian,yoon2018ganite,shalit2017estimating,zhang2020learning}.
	The global model interpretation method in \cite{yang2018global} can analyze the subgroup structure of arbitrary models but it depends on local model interpreters and does not consider the treatment effects.

	For recursive partitioning, various criteria have been proposed.
	In the traditional RT, the criterion based on the mean squared error between the sample mean estimations from the training samples and the test samples is used \cite{johansson2018interpretable}, and it is referred to as the adaptive criterion.
	Based on the adaptive criterion, an honest criterion is proposed in \cite{athey2016recursive} by splitting the training samples into the training set and the estimation set to eliminate the bias of the adaptive criterion.
	In addition, a generalization cost is introduced to the adaptive or honest criterion in \cite{tran2019learning} to encourage generalization of the analysis.
	The interaction measure between the treatment and covariates is used as a partitioning criterion in \cite{su2009subgroup} and the parameter instability of the parametric models is used in \cite{seibold2016model}.
	In \cite{yang2018global}, the contribution matrix of the samples from local model interpreters is used.
	Some of these criteria implicitly consider the confidence of the estimation by the variance, but most of them do not provide a valid confidence interval of the estimation for each subgroup.
	In \cite{johansson2018interpretable}, a conformal regression method based on regression trees that provides the confidence interval is proposed, but the adaptive criterion is used for partitioning without any consideration of the confidence interval.

	\vspace{-0.1in}
	\section{Experiments} \vspace{-6pt}
	In this section, we evaluate R2P-HTE by comparing its performance with state-of-the-art HTE methods.
	Specifically, we compare R2P-HTE with four baselines:
	standard regression trees for causal effects (CT-A) \cite{breiman1984classification}, conformal regression trees for causal effects (CCT) \cite{johansson2018interpretable}, causal trees with honest criterion (CT-H) \cite{athey2016recursive}, and causal trees with generalization costs (CT-L) \cite{tran2019learning}.
	We implement CCT and CT-A by modifying the conformal regression tree and conventional regression tree methods for causal effects.  
	Details of the baseline algorithms are provided in the supplementary material.
	For the ITE estimator of R2P-HTE, here abbreviate as R2P, we use the causal multi-task Gaussian process (CMGP) \cite{alaa2017bayesian}.
	Because individual ground-truth treatment effects can never be observed in real data, we use two synthetic and two semi-synthetic datasets.  The first synthetic dataset (Synthetic dataset A) is the simple one proposed in \cite{athey2016recursive}.  Because  dataset A possesses little of the homogeneity within subgroups that is often found in the real world, we offer a second synthetic dataset B that possesses  greater homogeneity within subgroups and greater heterogenity across subgroups  and has many more features than A.  B was inspired by the initial clinical trial of remdesivir \cite{wang2020remdesivir} as a treatment for COVID-19 and uses the patient features listed in that trial, but not the data. In specific, it represents a discovery that remdesivir results in a faster time to clinical improvement for the patients with the shorter time from symptom onset to starting trial. 
	Aside from inspiration and features, B is relevant to COVID-19 only in that the COVID-19 is known to be a disease which displays in very heterogeneous ways.  The two semi-synthetic datasets are based on real world data; the first uses the Infant Health and Development Program (IHDP) dataset \cite{hill2011bayesian} and the second uses the Collaborative Perinatal Project (CPP) dataset \cite{dorie2019automated}.  Details of all these datasets are provided in the supplementary material.
	For each experiment, we conduct 50 simulations.  
	
The ``optimal'' ground-truth of subgroups depends on multiple objectives, including homogeneity, heterogeneity, and the number of subgroups. In the literature, a commonly adopted metric is the variance, because greater heterogeneity across subgroups and homogeneity within each subgroup generally imply well-discriminated subgroups.
	We denote the set of test samples by $\calD^{te}$ and the test samples that belong to subgroup $l$ as $\calD^{te}_l$.  We define the mean and variance of the treatment effects of the test samples in subgroup $l$ as $\textrm{Mean}(\calD^{te}_l)$ and $\textrm{Var}(\calD^{te}_l)$, respectively.
	We define {\em heterogeneity across  subgroups} to be the variance of the mean of the treatment effects: 
	$V^{\textrm{across}}=\textrm{Var}(\{\textrm{Mean}(\calD^{te}_l)\}_{l=1}^{L})$, where $L$ is the number of subgroups; we define the {\em average in-subgroup variance} to be 
	$
	V^{\textrm{in}}(\calD^{te})=\frac{1}{L}\sum_{l=1}^L \textrm{Var}(\calD^{te}_l)
	$.
	We also provide  the average number of subgroups for better understanding of the results.
	We set the miscoverage rate to be $\alpha=0.05$, so we demand a 95\% ITE coverage rate.  (We do not report actual coverage rates here because all methods achieve the target coverage rate, but they are reported in the supplementary material, along with other details.)

	\begin{table}[ht]
		\caption{\textbf{Comparison of methods}~~ For the measures $V^{\textrm{across}}$ and $V^{\textrm{in}}$, and for the widths of confidence intervals, we highlight the best results in bold.}
		\footnotesize
		\label{table:experiments}
		\begin{center}
			\setlength\tabcolsep{3pt}
			\begin{tabular}{c|c|c|c|c|c|c|c|c}
				\toprule
				& \multicolumn{4}{c|}{Synthetic dataset A} & \multicolumn{4}{c}{Synthetic dataset B} \\
				\midrule
				& $V^{\textrm{across}}$ & $V^{\textrm{in}}$  & \# SGs   & CI width      & $V^{\textrm{across}}$ & $V^{\textrm{in}}$  & \# SGs   & CI width      \\
				\midrule
				\textsc{R2P}      & \textbf{0.22$\pm$.01} & \textbf{0.03$\pm$.001} & 4.9$\pm$.16 & \textbf{0.08$\pm$.003}
				& \textbf{2.39$\pm$.04} & \textbf{0.12$\pm$.01} & 5.0$\pm$.16 & \textbf{0.88$\pm$.06} \\
				\textsc{CCT} 	  & 0.18$\pm$.02 & 0.05$\pm$.01 & 4.4$\pm$.24 & 7.42$\pm$.48
				& 1.97$\pm$.14 & 0.58$\pm$.15 & 5.0$\pm$.13 & 5.95$\pm$.59 \\
				\textsc{CT-A}     & 0.19$\pm$.02 & 0.04$\pm$.01 & 4.7$\pm$.21 & 3.96$\pm$.16
				& 2.24$\pm$.06 & 0.30$\pm$.05 & 5.1$\pm$.15 & 2.77$\pm$.20  \\
				\textsc{CT-H}     & 0.12$\pm$.03 & 0.11$\pm$.02 & 3.1$\pm$.39 & 4.39$\pm$.22
				& 2.01$\pm$.13 & 0.53$\pm$.13 & 4.5$\pm$.15 & 3.38$\pm$.32  \\
				\textsc{CT-L}     & 0.12$\pm$.02 & 0.10$\pm$.02 & 2.9$\pm$.35 & 5.22$\pm$.02
				& 0.80$\pm$.26 & 1.77$\pm$.27 & 3.1$\pm$.28 & 6.92$\pm$.53 \\
				\midrule
				\midrule
				& \multicolumn{4}{c|}{IHDP dataset}& \multicolumn{4}{c}{CPP dataset} \\
				\midrule
				& $V^{\textrm{across}}$ & $V^{\textrm{in}}$  & \# SGs   & CI width     & $V^{\textrm{across}}$ & $V^{\textrm{in}}$  & \# SGs   & CI width     \\
				\midrule
				\textsc{R2P}      & \textbf{0.46$\pm$.04} & \textbf{0.38$\pm$.03} & 4.1$\pm$.12 & \textbf{1.27$\pm$.22}
				& \textbf{0.06$\pm$.02} & \textbf{0.10$\pm$.01} & 5.7$\pm$.30 & \textbf{1.11$\pm$.13}\\
				\textsc{CCT} 	  & 0.30$\pm$.04 & 0.43$\pm$.05 & 4.3$\pm$.13 & 5.70$\pm$.23
				& 0.03$\pm$.02 & 0.12$\pm$.01 & 6.4$\pm$.20 & 3.60$\pm$.12 \\
				\textsc{CT-A}     & 0.31$\pm$.04 & 0.57$\pm$.05 & 4.1$\pm$.08 & 3.71$\pm$.08
				& 0.03$\pm$.01 & 0.12$\pm$.01 & 6.6$\pm$.18 & 2.45$\pm$.06 \\
				\textsc{CT-H}     & 0.28$\pm$.05 & 0.56$\pm$.05 & 3.8$\pm$.14 & 3.76$\pm$.14
				& 0.01$\pm$.00 & 0.14$\pm$.01 & 5.2$\pm$.23 & 2.67$\pm$.06 \\
				\textsc{CT-L}     & 0.27$\pm$.06 & 0.64$\pm$.05 & 2.8$\pm$.23 & 4.75$\pm$.15
				& 0.01$\pm$.01 & 0.14$\pm$.01 & 2.9$\pm$.29 & 3.23$\pm$.07 \\
				\bottomrule
			\end{tabular}
		\end{center}
	\end{table}
	\vspace{-6pt}

\begin{figure}[t]
		\vspace{-0.5em}
		\centering
		\includegraphics[width=1.\linewidth]{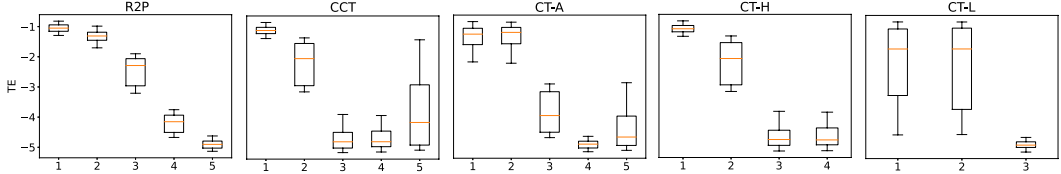}
		\caption{Treatment effects for subgroups identified by each algorithm when applied to Synthetic dataset B. Each box represents the range between the 25th and 75th percentiles of the treatment effects on the test samples; each whisker represents the range between the 5th and 95th percentiles.
			\label{fig:boxplot_synthetic}}
		\vspace{-1.5em}
	\end{figure}

	\begin{paragraph}{Results}
		Table \ref{table:experiments} reports the performances of R2P and the baselines for all four  datasets.  Keep in mind that {\em larger} $V^{\textrm{across}}$ means  greater heterogeneity  {\em across}  subgroups, while {\em smaller} $V^{\textrm{in}}$ means  greater  homogeneity {\em within}  subgroups.  
		As the table shows, R2P displays by far the best performance on all four datasets:  the greatest heterogeneity across  subgroups, the greatest homogeneity within subgroups, and the narrowest confidence intervals.  It accomplishes all this while identifying comparable numbers of subgroups.  We conclude that R2P identifies subgroups more effectively than any of the other methods, and allows fewer false discoveries (as was illustrated in the Introduction).  The performance  of R2P reflects one of its strengths: the ability to use {\em any} method of estimating ITE.
		
		The  effectiveness of R2P can also be seen in Fig. \ref{fig:boxplot_synthetic}, which provides, for R2P and each of the four baseline algorithms, boxplots of the distribution of the treatment effects for each identified subgroup for Synthetic dataset B.  We see that that R2P identifies subgroups reliably: different subgroups have very different average treatment effects and their distributions are non-overlapping or well-discriminated.  All the other methods are unreliable: false discovery  occurs for all four baseline methods, and occurs frequently for three of the four.  
	\end{paragraph}

	\vspace{-6pt}
	\begin{paragraph}{Gain from subgroup analysis}
		
		\begin{wraptable}{r}{7.1cm}
			\caption{Normalized $V^{\textrm{in}}$ of R2P\label{table:gain_subgroup}}
			\small
			\centering
			\setlength\tabcolsep{3pt}
			\begin{tabular}{c|c|c|c}
				\toprule  
				Synth A & Synth B & IHDP & CPP \\
				\midrule
				0.110$\pm$.005 & 0.046$\pm$.003 & 0.459$\pm$.033 & 0.691$\pm$.076 \\
				\bottomrule
			\end{tabular}
		\end{wraptable}
		To indicate the gain from subgroup analysis obtained by R2P, and hence to indicate the effectiveness of recursive partitioning, we compare $V^{\textrm{in}}$, the homogeneity within subgroups obtained by  R2P in Table \ref{table:experiments}, against the homogeneity within the entire population, $V^{\textrm{pop}}$. We divide $V^{\textrm{in}}$ by $V^{\textrm{pop}}$ to obtain the normalized $V^{\textrm{in}}$ in Table \ref{table:gain_subgroup}. Subgroup analysis with R2P reduces the average in-subgroup variance by 89\% and more than 95\% on Synthetic dataset A and B, respectively. For the semi-synthetic datasets, it reduces the average in-subgroup variance by more than 50\% and 30\% on the IHDP and CPP datasets, respectively. (Keep in mind that R2P constructs subgroups in a way that produces {\em both} strong heterogeneity across subgroups and strong homogeneity within subgroups.) 

	\end{paragraph}

	\vspace{-0.1in}
	\section{Conclusion} \vspace{-6pt}
	In this paper, we have studied robust HTE analysis based on recursive partitioning.  The algorithm proposed, R2P-HTE,  
	recursively partitions the entire population by taking into account both  heterogeneity across subgroups and homogeneity within subgroups, using the novel criterion of confident homogeneity that is based on the  quantification of uncertainty of the ITE estimation.  R2P-HTE robustly constructs subgroups and also provides confidence guarantees  for each subgroup.  Experiments for synthetic and semi-synthetic datasets (the latter based on real data) demonstrate that R2P-HTE outperforms state-of-the-art baseline algorithms in every dimension: greater heterogeneity across subgroups, greater homogeneity within subgroups, and  narrower confidence intervals.  One of the strengths of R2P-HTE is that it can employ {\em any} method for interpretable ITE estimation, including improved methods that will undoubtedly be developed in the future.

	\section*{Broader Impact} The understanding of treatment effects plays an important role in many areas, and especially in medicine and public policy.  In both areas, it is often the case that the same treatment has different effects on different groups; hence subgroup analysis is called for.  In medicine, subgroup analysis may make it possible to identify groups of patients (defined by covariates such as age, body mass index, blood pressure, etc.) suffering from a particular disease for whom a particular drug is effective and safe and other groups for whom the same drug is ineffective and unsafe.  Similarly, subgroup analysis may make it possible to identify groups of patients for whom one course of treatment (e.g. a particular mode of radiotherapy or chemotherapy) is preferable (more likely to be successful with fewer side effects) to another.  In public policy, subgroup analysis may make it possible to identify groups of people or geographic regions for which particular interventions (e.g., providing mosquito nets to combat malaria) are likely to be successful or unsuccessful.  The method for subgroup analysis that is developed in this paper, R2P, is an enormous improvement over state-of-the-art methods and therefore has the potential to make  enormous and widespread impact.  Moreover, because R2P can make use of improvements in the underlying estimation methods, this impact may grow over time.

		\section*{Acknowledgments}
		This work was supported by GlaxoSmithKline (GSK), the US Office of Naval Research (ONR), and the National Science Foundation (NSF): grant numbers 1407712, 1462245, 1524417, 1533983, 1722516. CS acknowledges the funding support from Kneron, Inc. We thank all reviewers for their comments and suggestions.

\bibliography{mybib}
\bibliographystyle{unsrt}

\clearpage
  \vbox{%
    \hsize\textwidth
    \linewidth\hsize
    \vskip 0.1in
  \hrule height 4pt
  \vskip 0.25in
  \vskip -\parskip%
    \centering
    {\LARGE\bf Supplementary Material for Robust Recursive Partitioning for Heterogeneous Treatment Effects with Uncertainty Quantification\par}
  \vskip 0.29in
  \vskip -\parskip
  \hrule height 1pt
  \vskip 0.09in%
  }

\renewcommand\thesection{\Alph{section}}
\setcounter{section}{0}

	\section{Preliminaries of Conformal Prediction}
	Here, we provide a basic idea of conformal prediction to help understanding.
	To this end, we introduce the following example in \cite{lei2018distribution}.
	Let $z_1, ...,z_n,z$ be samples of a scalar random variable exchangeably drawn from a distribution $P_Z$ and $z_{(1)},...,z_{(n)}$ denote the order statistics of $z_1,...,z_n$.
	We denote the $1-\alpha$-th quantile of $z_{(1)},...,z_{(n)}$ as
	\[
	\hat{Q}_{1-\alpha}=
	\begin{cases}
		z_{(\lceil(n+1)(1-\alpha)\rceil},& \text{if } \lceil(n+1)(1-\alpha)\rceil \leq n \\
		\infty,              & \text{otherwise}
	\end{cases}
	\]
	The rank of $z$ among $z_i,...,z$ is uniformly distributed over the set $\{1,...,n+1\}$ under the exchangeability assumption that the joint distribution of $z_1,...,z_n,z$ is invariant of the sampling order $z_i,...,z$.
	Thus, for a given miscoverage level $\alpha\in[0,1]$, we have $\mathbb{P}[z\leq\hat{Q}_{1-\alpha}]\leq 1-\alpha$ by summing the uniform distribution up to $\hat{Q}_{1-\alpha}$.
	
	This idea can be used in a regression problem with covariates $\bx\in\mathcal{X}$, where $d$ is the dimension of the covariate vector, and outcomes $y\in\mathcal{Y}$.
	Specifically, with the regressor $\hat{\mu}$, the confidence interval for $y$ is given as
	\[
	\hat{C}(\bx)=\left[ \hat{\mu}(\bx)-G_{1-\alpha}^{-1},\hat{\mu}(\bx)+G_{1-\alpha}^{-1} \right],
	\]
	where $G$ is the empirical distribution of the fitted residuals on the training samples (i.e., $|y_i-\hat{\mu}(\bx_i)|$, $i=1,...,n$) and $G_{1-\alpha}^{-1}$ is the $1-\alpha$-th quantile of $G$.
	However, this method may undercover $y$ since the residuals on the training samples typically smaller than those on the test samples due to overfitting.
	To avoid this, Split Conformal Regression (SCR) is introduced which separates the samples for training and computing the residuals.
	In SCP, the training samples is split into two equal-size subsets $\calI_1$ and $\calI_2$, and one subset $\calI_1$ is used to fit the regressor $\hat{\mu}_{\calI_1}$ and another one $\calI_2$ is used to compute the residuals for $\hat{\mu}_{\calI_1}$, $R_{\calI_2}=\{|y_i-\hat{\mu}_{\calI_1}(\bx_i)|:(\bx_i,y_i)\in\calI_2\}$.
	Based on the regressor and residuals, the confidence interval for $y$ with the regressor $\hat{\mu}_{\calI_1}$ is given as
	\[
	\hat{C}_{SCR}=\left[ \hat{\mu}_{\calI_1}(\bx)-\hat{Q}^{\calI_2}_{1-\alpha}, \hat{\mu}_{\calI_1}(\bx)+\hat{Q}^{\calI_2}_{1-\alpha} \right],
	\]
	where $\hat{Q}^{\calI_2}_{1-\alpha}$ is defined to be $(1-\alpha)(1+1/|\calI_2|)$-th quantile of $R_{\calI_2}$ (i.e., $\lceil (|\calI_2|+1)(1-\alpha)\rceil$-the smallest residual on $\calI_2$).
	Under the only one assumption of the exchangeability of the training samples $\{(\bx_i,y_i)\}^n_{i=1}$ and the testing sample $(\bx,y)$, it satisfies the following theorem.
	\begin{theorem}
		\cite{lei2018distribution}
		If the samples $\{(\bx_i,y_i)\}_{i=1}^n$ are exchangeable, then for a new sample $(\bx_{n+1},y_{n+1})$ drawn from $\calP_{X,Y}$,
		\[
		\mathbb{P}[y\in\hat{C}_{SCR}(\bx)]\geq 1-\alpha.
		\]
	\end{theorem}

	\section{Proof of Theorem 1}
	For subgroup $l$, let $\calP_{X,Y}^l$ be the distribution on $l\times\calY$, which is the conditional distribution $\calP_{X,Y|X\in l}$.
	According to Algorithm 1, we have the samples of subgroup $l$ from the entire samples consisting of two disjoint subsets as $\calD_l=\calI_1^l\cup\calI_2^l$, where $\calI_1^l$ is the samples that are used for training the estimator $\hat{\mu}_\calX$ and $\calI_2^l$ is the samples that are used for constructing the confidence interval $\hat{C}_l$.
	The samples in $\calD_l$ are exchangeable since they are i.i.d.
	Thus, for a new sample $(\bx_{n+1},y_{n+1})$ drawn from $\calP_{X,Y}^l$, we have $\mathbb{P}[y_{n+1}\in\hat{C}^{l}]\leq 1-\alpha$ from Theorem A.1.

	\section{Proof of Theorem 2}
	We have the samples of subgroup $l$ from the entire samples consisting of two disjoint subsets as $\calD_{\textrm{HTE},l}=\calI_1^l\cup\calI_2^l$. For training the estimators $\hat{\mu}^1_\calX$ and $\hat{\mu}^0_\calX$, the samples in $\calI_1^l$ are used. Also, the samples in $\calI_2^l$ are used to construct the confidence intervals $\hat{C}_l^1$ and $\hat{C}_l^0$ with the miscoverage rate $\sqrt{1-\alpha}$ according to the treatment indicator of the samples. Since the samples are i.i.d., for a new sample with each potential outcome, the corresponding confidence interval satisfies the miscoverage rate $\sqrt{1-\alpha}$ from Theorem A.1 as 
	\[
	\mathbb{P}[Y(1)\in\hat{C}^1_l(\bx)]\geq \sqrt{1-\alpha}\textrm{ and }\mathbb{P}[Y(0)\in\hat{C}^0_l(\bx)]\geq \sqrt{1-\alpha}.
	\]
	For the definitions of the ITE estimation $\hat{\tau}_l(\bx)$ and $\hat{C}_l^\tau(\bx)$, the events that the potential outcomes and ITE estimation belong to their corresponding confidence intervals satisfy the following relation:
	\[
	\{Y(1)\in\hat{C}^1_l(\bx)\}\cap\{Y(0)\in\hat{C}^0_l(\bx)\}\subset\{\tau\in\hat{C}^\tau_l(\bx)\},
	\]
	which implies that
	\[
	\mathbb{P}[\tau\in\hat{C}^\tau_l(\bx)]\geq \mathbb{P}[Y(1)\in\hat{C}^1_l(\bx)]\times\mathbb{P}[Y(0)\in\hat{C}^0_l(\bx)] \geq 1-\alpha.
	\]

	\section{Related Works in Subgroup Analysis with Recursive Partitioning}

	Subgroup analysis methods with recursive partitioning have been widely studied based on regression trees (RT) \cite{athey2016recursive,su2009subgroup,tran2019learning,johansson2018interpretable}.
	In these methods, the subgroups (i.e., leaves in the tree structure) are constructed; the treatment effects are estimated by the corresponding sample mean estimator on the leaf of the given covariates.
	To represent the non-linearity such as interactions between treatment and covariates \cite{doove2014comparison},
	a parametric model is integrated into regression trees for subgroup analysis \cite{seibold2016model}.
	However, such approach can be used only for the limited types of models, which is not particularly satisfying given the fact that advanced causal inference models based on deep neural networks or multi-task Gaussian processes have been studied which outperform the traditional estimators \cite{alaa2017bayesian,yoon2018ganite,shalit2017estimating,zhang2020learning}.
	The global model interpretation method in \cite{yang2018global} can analyze the subgroup structure of arbitrary models but it depends on local model interpreters and does not consider the treatment effects.

	\begin{table}[ht]
		\caption{Comparison of related works}
		\small
		\label{table:trials}
		\begin{center}
			\setlength\tabcolsep{3pt}
			\begin{tabular}{cccccc}
				\toprule
				\multirow{2}{*}{\textbf{Method}} & \multirow{2}{*}{\textbf{Estimator}} & \multicolumn{3}{c}{\textbf{Partitioning Criterion}} & \multirow{2}{*}{\textbf{Treatment effect}} \\
				\cmidrule{3-5}
				&  & \textbf{Type} & \textbf{Homogeneity} & \textbf{CI widths} &  \\
				\midrule
				\cite{athey2016recursive} & Sample mean & Honest criterion & $\times$  & $\times$ & $\surd$ \\
				\midrule
				\cite{tran2019learning} & Sample mean & \makecell{Adaptive criterion with\\ generalization cost} & $\times$  & $\times$ & $\surd$ \\
				\midrule
				\cite{johansson2018interpretable} & Sample mean & Adaptive criterion & $\times$  & $\times$ & $\times$ \\
				\midrule
				\cite{su2009subgroup} & Sample mean & Interaction measure & $\times$  & $\times$ & $\surd$ \\
				\midrule
				\cite{seibold2016model} & Parametric model & Parameter instability & $\times$  & $\times$ & $\surd$ \\
				\midrule
				\cite{yang2018global} & Arbitrary estimator & Feature contribution & $\times$  & $\times$ & $\times$ \\
				\midrule
				\midrule
				Our work & Arbitrary estimator & \makecell{Confident criterion} & $\surd$  & $\surd$ & $\surd$ \\
				\bottomrule
			\end{tabular}
		\end{center}
	\end{table}

	For recursive partitioning, various criteria have been proposed.
	In the traditional RT, the criterion based on the mean squared error of the estimated means on the training samples and the test samples is used \cite{johansson2018interpretable}, and it is referred to as the adaptive criterion.
	Basically, this adaptive criterion identifies subgroups with heterogeneous treatment effects by trying to maximize the heterogeneity across the identified subgroups.
	Based on the adaptive criterion, in \cite{athey2016recursive}, an honest criterion is proposed.
	In the criterion, the training samples are split into two subsets;
	One is used to build a tree structure and another one is used to estimate the treatment effects.
	By doing this, the honest criterion can eliminate the bias of the adaptive criterion.
	In \cite{tran2019learning}, a generalization cost is introduced to encourage generalization of the analysis.
	It is defined by using another subset of the training samples and adopted to
	the adaptive or honest criterion.
	The interaction measure between the treatment and covariates is used as a partitioning criterion in \cite{su2009subgroup} and the parameter instability of the parametric models is used in \cite{seibold2016model}.
	In \cite{yang2018global}, the contribution matrix of the samples from local model interpreters is used for partitioning.
	These criteria focus on the heterogeneity across the subgroups, but the variant of the treatment effects within each subgroup (i.e., the homogeneity within each subgroup) is neglected.
	Besides, some of these criteria construct the confidence intervals using the estimated variances,  but most of these intervals fail to achieve the coverage guarantee for each subgroup in finite samples. In \cite{johansson2018interpretable}, a conformal regression method for constructing confidence intervals using regression trees is proposed. The adaptive criterion they use for partitioning does not take into account the confidence interval. The confident criterion in R2P is different from these criteria by
	considering both heterogeneity and homogeneity of subgroups and constructing subgroups with confidence intervals.

	\section{Description of Datasets}
	\subsection{Description of Synthetic Models}
	\begin{paragraph}{Synthetic dataset A}
		We first consider the synthetic treatment effect model proposed in \cite{athey2016recursive}.
		It describes the potential outcome $y$ for given treatment $t\in\{0,1\}$ as follows:
		\[
		y_i(t)=\eta(\bx_i)+\frac{1}{2}(2t-1)\kappa(\bx_i)+\epsilon_i,
		\]
		where $\epsilon_i\sim \mathcal{N}(0,0.01)$, $x_{i,k}\sim\mathcal{N}(0,1)$, and $\eta(\cdot)$ and $\kappa(\cdot)$ are the design functions.
		The response surface of outcome $y_i$'s is determined by the design functions.
		The functions $\eta(\cdot)$ and $\kappa(\cdot)$ are the mean outcome and treatment effect for some given covariates, respectively. In this synthetic model, we consider the following design with 2-dimensional covariates, $\eta(\bx)=\frac{1}{2}x_1+x_2$ and $\kappa(\bx)=\frac{1}{2}x_1$.
		There is no redundant covariate which has no effect on the outcomes.
		In the experiments, we generate 300 samples for training and 1000 samples for testing.
	\end{paragraph}
	
	\begin{paragraph}{Synthetic dataset B}
		Here, we introduce a synthetic model based on the initial clinical trial results of remdesivir to COVID-19 \cite{wang2020remdesivir}. The result shows that remdesivir results in a faster time to clinical improvement for the patients with a shorter time from symptom onset to starting the trial.
		Since the clinical trial data is not public, we generate a synthetic model in which the treatment effects mainly depends on the time from symptom onset to trial based on the clinical trial setting and results in the paper \cite{wang2020remdesivir}.
		We consider the following 10 baseline covariates:
		age $\sim\mathcal{N}(66,4)$,
		white blood cell count ($\times 10^9$ per L) $\sim\mathcal{N}(66,4)$,
		lymphocyte count ($\times10^9$ per L) $\sim\mathcal{N}(0.8,0.1)$,
		platelet count ($\times10^9$ per L) $\sim\mathcal{N}(183,20.4)$,
		serum creatinine (U/L) $\sim\mathcal{N}(68,6.6)$,
		asparatate aminotransferase (U/L) $\sim\mathcal{N}(31,5.1)$,
		alanine aminotransferase (U/L) $\sim\mathcal{N}(26,5.1)$,
		lactate dehydrogenase (U/L) $\sim\mathcal{N}(339,51)$,
		creatine kinase (U/L) $\sim\mathcal{N}(76,21)$, and
		time from symptom onset to starting the trial (days) $\sim\textrm{Unif}(4,14)$.
		We approximate the distribution using the patient characteristics provided in the paper.
		To construct treatment/control responses, we first adopt the response surface in the IHDP dataset \cite{hill2011bayesian} for the covariates except for the time.
		We then use a logistic function on the time covariate to produce different effectiveness (i.e., the faster time to clinical improvement with the shorter time from symptom onset to the trial).
		Specifically, the control response is defined as 
		$Y(0)\sim\mathcal{N}(X_{-0}\boldsymbol{\upbeta}+(1+e^{-(x_{0}-9)})^{-1}+5,0.1)$, and the treated response is defined as
		$Y(1)\sim\mathcal{N}(X_{-0}\boldsymbol{\upbeta}+5\cdot(1+e^{-(x_{0}-9)})^{-1},0.1)$,
		where $X_{-0}$ represents the matrix of the standardized (zero-mean and unit standard deviation) covariate values except for the time covariate $x_{0}$ and the coefficients in the vector $\boldsymbol{\upbeta}$ are randomly sampled among the values $(0,0.1,0.2,0.3,0.4)$ with the probability $(0.6,0.1,0.1,0.1,0.1)$, respectively. In this synthetic model,
		the response surface is consistent with the trial result in \cite{wang2020remdesivir} such that the time to clinical improvement (i.e., the treatment effect) becomes faster as the shorter time from symptom onset to the trial.
	\end{paragraph}
	
	\subsection{Description of Semi-Synthetic Datasets}
	We consider two semi-synthetic datasets for treatment effect estimation: the Infant Health and Development Program (IHDP) \cite{hill2011bayesian} and the Collaborative Perinatal Project (CPP) \cite{dorie2019automated}.
	
	\begin{paragraph}{IHDP dataset}
		The Infant Health and Development Program (IHDP) is a randomized experiment intended to enhance the cognitive and health status of low-birth-weight, premature infants through intensive high-quality child care and home visits from a trained provider.
		Based on the real experimental data about the impact of the IHDP on the subjects' IQ scores at the age of three,
		the semi-synthetic (simulated) dataset is developed and has been used to evaluate treatment effects estimation in \cite{hill2011bayesian,alaa2017bayesian,shalit2017estimating,louizos2017causal}.
		All outcomes (i.e., response surfaces) are simulated using the real covariates.
		The dataset consists of 747 subjects (608 untreated and 139 treated), and 25 input covariates for each subject.
		We generated the outcomes using the standard non-linear mean outcomes of ``Response Surface B'' setting provided in \cite{hill2011bayesian}. A noise $\epsilon\sim\mathcal{N}(0,0.1)$ is added to each observed outcome. In the experiments, we use 80\% samples for training and 20\% samples for testing.
	\end{paragraph}
	
	\begin{paragraph}{CPP dataset}
		In the 2016 Atlantic Causal Inference Conference competition (ACIC),
		a semi-synthetic dataset is developed based on the data from the Collaborative Perinatal Project (CPP) \cite{dorie2019automated}.
		It comprises of multiple datasets that are generated by distinct data generating processes (causal graphs) and random seeds.
		Each dataset consists of 4802 observations with 58 covariates of which 3 are categorical, 5 are binary, 27 are count data, and the remaining 23 are continuous. The factual and counterfactual samples are drawn from a generative model and a noise $\epsilon\sim\mathcal{N}(0,0.1)$ is added for each observed outcome.
		In the experiments, we use the dataset with index 1 provided in \cite{dorie2019automated} and drop the rows whose $Y(1)$ or $Y(0)$ above the 99\% quantile or below the 1\% quantile to avoid the outliers.
		The dataset consists of 35\% treated subjects and 65\% untreated subjects.
		We randomly pick 500 samples for training and 300 samples for testing from the dataset.
	\end{paragraph}

	\section{Description of Algorithms in Experiments}
	
	\begin{paragraph}{Robust recursive partitioning for HTE (R2P-HTE)}
		We implement R2P-HTE based on Section 4 of the main manuscript.
		For the ITE estimator, we use the causal multi-task Gaussian process (CMGP) in \cite{alaa2017bayesian}.
		Using the outcome estimates from CMGP (i.e., $\hat{\mu}^1(\bx)$ and $\hat{\mu}^0(\bx)$), we construct the confidence interval for the ITE estimator $\hat{\tau}(\bx)$.
		Then, we can apply Algorithm 1 in the main manuscript.
		In the experiments, we set $\lambda=0.5$ and $\gamma=0.05$.
		(We use $\lambda=0$ for the CPP dataset to avoid the excessive effect of the confidence intervals compared with the heterogeneity effect in the dataset.)
		We use $\alpha=0.1$ and 0.5 split ratio for split conformal prediction.
		We set the minimum number of training samples in each leaf as 10.
		To avoid excessive conservativeness, we use the confidence interval of the miscoverage rate $\beta=0.8$ for the expected deviation, $\hat{S}_l$, in the confident split criterion.
		We set the hyper-parameters manually as above considering a typical value (e.g., 0.95 is a typical value for significance tests in tree-based methods), but if needed, the hyper-parameter tuning can be done by a grid search method as in a typical recursive partitioning methods \cite{tran2019learning}.
	\end{paragraph}

	\begin{paragraph}{Standard regression tree for causal effects (CT-A)}
		Because a standard regression tree in \cite{breiman1984classification} is not developed for estimating treatment effects, we implement the modified version of the standard regression tree for causal effects estimation in \cite{athey2016recursive}.
		In this modified version, the regression tree recursively partitions according to a criterion based on the expectation of the mean squared error (MSE) of the treatment effects.
		In the literature, this criterion called an adaptive criterion.
		We refer to \cite{athey2016recursive} for more details of the method.
		In the experiments, we set the minimum number of training samples in each leaf as 20 since CT-A does not need to split the data samples into two subsets for validation as in other methods.
		After building the tree-structure, we prune the tree according to the statistical significance gain at 0.05.
	\end{paragraph}

	\begin{paragraph}{Conformal regression tree for causal effects (CCT)}
		We modify the conformal regression tree \cite{johansson2018interpretable} for our experiments of treatment effect estimation. We implemented CCT by applying the split conformal prediction method to a standard causal tree (i.e., CT-A). The ITE confidence interval is constructed in the same way as R2P.
		In the experiments, we use 0.5 split ratio for the split conformal prediction method and set the minimum number of training samples in each leaf as 10.
		After building the tree-structure, we prune the tree according to the statistical significance gain at 0.05.
	\end{paragraph}

	\begin{paragraph}{Causal tree with honest criterion (CT-H)}
		We implement a causal tree method proposed in \cite{athey2016recursive}.
		The method modify the standard regression tree for causal effects in which an honest criterion is used instead of the adaptive criterion.
		It divides tree-building and treatment effect estimation into two steps.
		The samples are split into two subsets: training samples to build the tree and samples to estimate treatment effects. This two-step procedure makes the tree-building and the treatment effect estimation process independent, which can eliminate the bias in treatment effect estimation.
		We refer to \cite{athey2016recursive} for more details of the method,
		In the experiments, we use 0.5 split ratio for building the tree  and estimating the effects. We set the minimum number of training samples in each leaf as 10.
		After building the tree-structure, we prune the tree according to the statistical significance gain at 0.05.
	\end{paragraph}
	
	\begin{paragraph}{Causal tree with generalization costs (CT-L)}
		We implement a causal tree with a criterion considering generalization costs in \cite{tran2019learning}.
		This method is a modified version of the causal tree in \cite{athey2016recursive}.
		It splits the data samples into the training and validation samples, and builds the tree using the training samples while penalizing based on generalization ability using the validation samples.
		For more details of the method, we refer to \cite{tran2019learning}.
		In the experiments, we use 0.5 split ratio for the training and validation. We set the minimum number of training samples in each leaf as 10.
		After building the tree-structure, we prune the tree according to the statistical significance gain at 0.05.
	\end{paragraph}

	\section{Additional Experimental Results}
	\subsection{Average Overlap of Treatment Effects across Subgroups}

	\newlength{\oldintextsep}
	\setlength{\oldintextsep}{\intextsep}
	\setlength\intextsep{0pt}
	\begin{wraptable}{r}{7.1cm}
		\caption{Average overlap of treatment effects across subgroups.}
		\footnotesize
		\begin{center}
			\setlength\tabcolsep{4pt}
			\vspace{-3pt}
			\begin{tabular}{c|c|c|c|c}
				\toprule
				& Synth A & Synth B & IHDP & CPP \\
				\midrule
				R2P      & \textbf{0.45$\pm$.06} & \textbf{0.14$\pm$.03} & \textbf{0.32$\pm$.04} & \textbf{0.23$\pm$.03} \\
				CCT		 & 1.35$\pm$.04 & 0.63$\pm$.15 & 0.81$\pm$.09 & 0.55$\pm$.04 \\
				CT-A     & 1.13$\pm$.21 & 0.44$\pm$.09 & 0.59$\pm$.08 & 0.47$\pm$.05 \\
				CT-H     & 0.60$\pm$.20 & 0.60$\pm$.16 & 0.76$\pm$.10 & 0.45$\pm$.05 \\
				CT-L     & 0.87$\pm$.18 & 2.27$\pm$.55 & 0.46$\pm$.10 & 0.24$\pm$.04 \\
				\bottomrule
			\end{tabular}
		\end{center}
		\vskip -0.1in
	\end{wraptable}
	As one metric for evaluating false discovery, we can use the overlap of treatment effects across subgroups in Fig. 3 of the main manuscript. The average overlap of treatment effects across subgroups indicates whether the
	subgroups are well-discriminated. Specifically, we define a treatment effect interval of each subgroup $l$ as $[a_l(p),b_l(q)]$, where $a_l(p)$ and $b_l(q)$ are $p$-th and $q$-th percentiles of the treatment effects in the subgroup $l$. We define the average overlap of treatment effects across subgroups as the overlapped width of the treatment effect intervals between all the pairs of the subgroups.
	We provide the average overlaps of R2P and the baselines for all datasets with $p=20$ and $q=80$.
	The table shows that the average overlap width of R2P is significantly small than the one of the baselines, which implies that R2P performs best for discriminating the subgroups.

	\subsection{Results with Maximum Depth for Partitioning}
	We provide the results of the maximum depth for partitioning in R2P to demonstrate the effectiveness of the confident criterion more clearly.
	We set the maximum depth of each method to be 2, which limits the maximum number of identified subgroups by 4.
	In most datasets, R2P has both the highest variance across subgroups and lowest in-subgroup variance.
	This implies that each partitioning in R2P is more effective to identify the subgroups than that in the other methods.
	In Synthetic dataset B, CT-A has the highest variance across subgroups, but its difference from that of R2P is marginal and the in-subgroup variance of CT-A is much higher than that of R2P.

	\begin{table}[ht]
		\caption{Results with maximum depth for partitioning.}
		\fontsize{8pt}{8pt}\selectfont
		\begin{center}
			\setlength\tabcolsep{2pt}
			\begin{tabular}{c|c|c|c|c|c|c|c|c|c|c}
				\toprule
				& \multicolumn{5}{c|}{Synthetic dataset A} 
				& \multicolumn{5}{c}{Synthetic dataset B} \\
				\midrule
				& $V^{\textrm{across}}$ & $V^{\textrm{in}}$  & \# SGs   & CI width & Cov. (\%)
				& $V^{\textrm{across}}$ & $V^{\textrm{in}}$  & \# SGs   & CI width & Cov. (\%)  \\
				\midrule
				\textsc{R2P}      & \textbf{0.27$\pm$.01} & \textbf{0.04$\pm$.001} & 4.0$\pm$.04 & \textbf{0.09$\pm$.002} & 99.06$\pm$.23
				& 2.19$\pm$.04 & \textbf{0.14$\pm$.01} & 4.0$\pm$.00 & \textbf{0.98$\pm$.07} & 99.28$\pm$.16 \\
				\textsc{CCT} 	  & 0.20$\pm$.02 & 0.07$\pm$.01 & 3.4$\pm$.22 & 8.28$\pm$.42 & 100.0$\pm$.00
				& 2.10$\pm$.08 & 0.43$\pm$.09 & 3.9$\pm$.09 & 6.05$\pm$.46  & 99.99$\pm$.01 \\
				\textsc{CT-A}     & 0.24$\pm$.02 & 0.06$\pm$.01 & 3.6$\pm$.15 & 4.29$\pm$.16  & 99.99$\pm$.02
				& \textbf{2.23$\pm$.06} & 0.30$\pm$.06 & 3.9$\pm$.08 & 2.97$\pm$.20  & 98.68$\pm$.52 \\
				\textsc{CT-H}     & 0.14$\pm$.03 & 0.11$\pm$.02 & 2.5$\pm$.27 & 4.71$\pm$.18  & 99.99$\pm$.01
				& 2.13$\pm$.09 & 0.45$\pm$.09 & 3.8$\pm$.12 & 3.45$\pm$.25  & 98.66$\pm$.68 \\
				\textsc{CT-L}     & 0.13$\pm$.03 & 0.12$\pm$.02 & 2.4$\pm$.25 & 5.49$\pm$.16 & 99.99$\pm$.01
				& 0.82$\pm$.23 & 1.74$\pm$.25 & 2.8$\pm$.19 & 6.71$\pm$.53  & 99.70$\pm$.30 \\
				\midrule
				\midrule
				& \multicolumn{5}{c|}{IHDP dataset}
				&  \multicolumn{5}{c}{CPP dataset} \\
				\midrule
				& $V^{\textrm{across}}$ & $V^{\textrm{in}}$  & \# SGs   & CI width & Cov. (\%)
				& $V^{\textrm{across}}$ & $V^{\textrm{in}}$  & \# SGs   & CI width & Cov. (\%)  \\
				\midrule
				\textsc{R2P}      & \textbf{0.52$\pm$.05} & \textbf{0.42$\pm$.05} & 3.5$\pm$.14 & \textbf{1.21$\pm$.13} & 97.24$\pm$.50
				& \textbf{0.05$\pm$.02} & \textbf{0.12$\pm$.01} & 3.7$\pm$.13 & \textbf{1.22$\pm$.14} & 99.04$\pm$.23\\
				\textsc{CCT} 	  & 0.28$\pm$.05 & 0.62$\pm$.06 & 3.5$\pm$.14 & 6.11$\pm$.21 & 99.55$\pm$.14
				& \textbf{0.05$\pm$.03} & 0.13$\pm$.01 & 3.4$\pm$.14 & 3.66$\pm$.13 & 99.50$\pm$.21 \\
				\textsc{CT-A}     & 0.33$\pm$.04 & 0.58$\pm$.05 & 3.7$\pm$.13 & 3.64$\pm$.08 & 97.25$\pm$.43
				& 0.02$\pm$.01 & 0.13$\pm$.01 & 3.5$\pm$.14 & 2.44$\pm$.05 & 96.17$\pm$.51 \\
				\textsc{CT-H}     & 0.30$\pm$.05 & 0.60$\pm$.05 & 3.5$\pm$.14 & 3.72$\pm$.12 & 97.17$\pm$.43
				& 0.01$\pm$.00 & 0.14$\pm$.01 & 3.4$\pm$.14 & 2.59$\pm$.06 & 97.31$\pm$.56 \\
				\textsc{CT-L}     & 0.30$\pm$.06 & 0.67$\pm$.04 & 2.7$\pm$.18 & 4.72$\pm$.17 & 98.99$\pm$.23
				& 0.02$\pm$.02 & 0.13$\pm$.01 & 2.7$\pm$.21 & 3.25$\pm$.07 & 99.55$\pm$.22 \\
				\bottomrule
			\end{tabular}
		\end{center}
	\end{table}

	\subsection{Results of R2Ps with Different ITE Estimators}
	
	Here, we provide the results of R2Ps with different ITE estimators in Table \ref{table:diff_ITE}.
	For this, we integrate R2P with the ITE estimators based on dragonnet (DN) \cite{shi2019adapting}, random forest (RF), and CT-H \cite{athey2016recursive}.
	To evaluate the precision of the ITE estimation, we introduce a precision in estimation of heterogeneous effect (PEHE)  defined as follows \cite{hill2011bayesian}:
	\[
	\textrm{PEHE}=\frac{1}{n}\sum_{i=1}^n ((\hat{\mu}^1(X_i)-\hat{\mu}^0(X_i))-\mathbb{E}[Y_i^{(1)}Y_i^{(1)}|X_i=\bx])^2.
	\]
	The lower PEHE implies a more accurate ITE estimation. In the table, we can see that the use of a better estimator allows R2P to construct better subgroups. This clearly shows that R2P can effectively exploit the better precision of the ITE estimation when constructing subgroups. Besides, R2P can seamlessly integrate any ITE estimators. Consequently, we can expect that R2P constructs better subgroups by using some improved ITE estimators in the future.

	\begin{table}[ht]
		\caption{Results of R2Ps with different ITE estimators.}
		\fontsize{8pt}{8pt}\selectfont
		\begin{center}
			\label{table:diff_ITE}
			\setlength\tabcolsep{2pt}
			\begin{tabular}{c|c|c|c|c|c|c|c|c|c|c}
				\toprule
				& \multicolumn{5}{c|}{Synthetic dataset A} 
				& \multicolumn{5}{c}{Synthetic dataset B} \\
				\midrule
				& $V^{\textrm{across}}$ & $V^{\textrm{in}}$  & \# SGs   & CI width & $\sqrt{\textrm{PEHE}}$ 
				& $V^{\textrm{across}}$ & $V^{\textrm{in}}$  & \# SGs   & CI width & $\sqrt{\textrm{PEHE}}$  \\
				\midrule
				\textsc{R2P}      & 0.22$\pm$.01 & 0.03$\pm$.001 & 4.9$\pm$.16 & 0.08$\pm$.003 & 0.01$\pm$.00
				& 2.39$\pm$.04 & 0.12$\pm$.01  & 5.0$\pm$.16 & 0.88$\pm$.06  & 0.16$\pm$.01 \\
				\textsc{R2P-DN}   & 0.21$\pm$.02 & 0.05$\pm$.01  & 4.9$\pm$.27 & 0.71$\pm$.05  & 0.06$\pm$.00
				& 2.32$\pm$.06 & 0.19$\pm$.03  & 5.1$\pm$.17 & 2.24$\pm$.09  & 0.41$\pm$.02 \\
				\textsc{R2P-RF}   & 0.18$\pm$.02 & 0.08$\pm$.01  & 4.9$\pm$.26 & 2.91$\pm$.12  & 0.33$\pm$.01
				& 2.20$\pm$.08 & 0.33$\pm$.07  & 5.1$\pm$.15 & 3.10$\pm$.32  & 0.42$\pm$.04 \\
				\textsc{R2P-CT-H} & 0.12$\pm$.02 & 0.07$\pm$.03  & 4.3$\pm$.49 & 8.87$\pm$.32  & 0.51$\pm$.02
				& 1.05$\pm$.25 & 1.51$\pm$.25  & 4.6$\pm$.41 & 6.42$\pm$.51  & 0.92$\pm$.12 \\
				\midrule
				\midrule	
				& \multicolumn{5}{c|}{IHDP dataset}
				&  \multicolumn{5}{c}{CPP dataset} \\
				\midrule
				& $V^{\textrm{across}}$ & $V^{\textrm{in}}$  & \# SGs   & CI width & $\sqrt{\textrm{PEHE}}$
				& $V^{\textrm{across}}$ & $V^{\textrm{in}}$  & \# SGs   & CI width & $\sqrt{\textrm{PEHE}}$  \\
				\midrule
				\textsc{R2P}      & 0.46$\pm$.04 & 0.38$\pm$.03 & 4.1$\pm$.12 & 1.27$\pm$.22 & 0.22$\pm$.02
				& 0.06$\pm$.02 & 0.10$\pm$.01 & 5.7$\pm$.30 & 1.11$\pm$.13 & 0.13$\pm$.01\\
				\textsc{R2P-DN}   & 0.41$\pm$.03 & 0.44$\pm$.04 & 4.3$\pm$.13 & 1.92$\pm$.09 & 0.33$\pm$.01
				& 0.06$\pm$.02 & 0.10$\pm$.01 & 6.0$\pm$.26 & 1.52$\pm$.07 & 0.19$\pm$.01 \\
				\textsc{R2P-RF}   & 0.32$\pm$.05 & 0.55$\pm$.05 & 4.0$\pm$.32 & 3.07$\pm$.13 & 0.39$\pm$.02
				& 0.04$\pm$.02 & 0.12$\pm$.01 & 6.0$\pm$.27 & 1.80$\pm$.06 & 0.23$\pm$.01 \\
				\textsc{R2P-CT-H} & 0.09$\pm$.04 & 0.75$\pm$.05 & 2.7$\pm$.50 & 6.56$\pm$.25 & 0.83$\pm$.03
				& 0.00$\pm$.00 & 0.14$\pm$.00 & 1.0$\pm$.004 & 4.20$\pm$.10 & 0.41$\pm$.01 \\
				\bottomrule
			\end{tabular}
		\end{center}
	\end{table}

	\subsection{Non-Interpretability of Grouping Using Quantiles of ITE Estimation}
	
	
	
\begin{figure}[ht]
		\vspace{0.1in}
		\centering
		\begin{subfigure}[b]{0.22\textwidth}
			\centering
			\includegraphics[width=\textwidth]{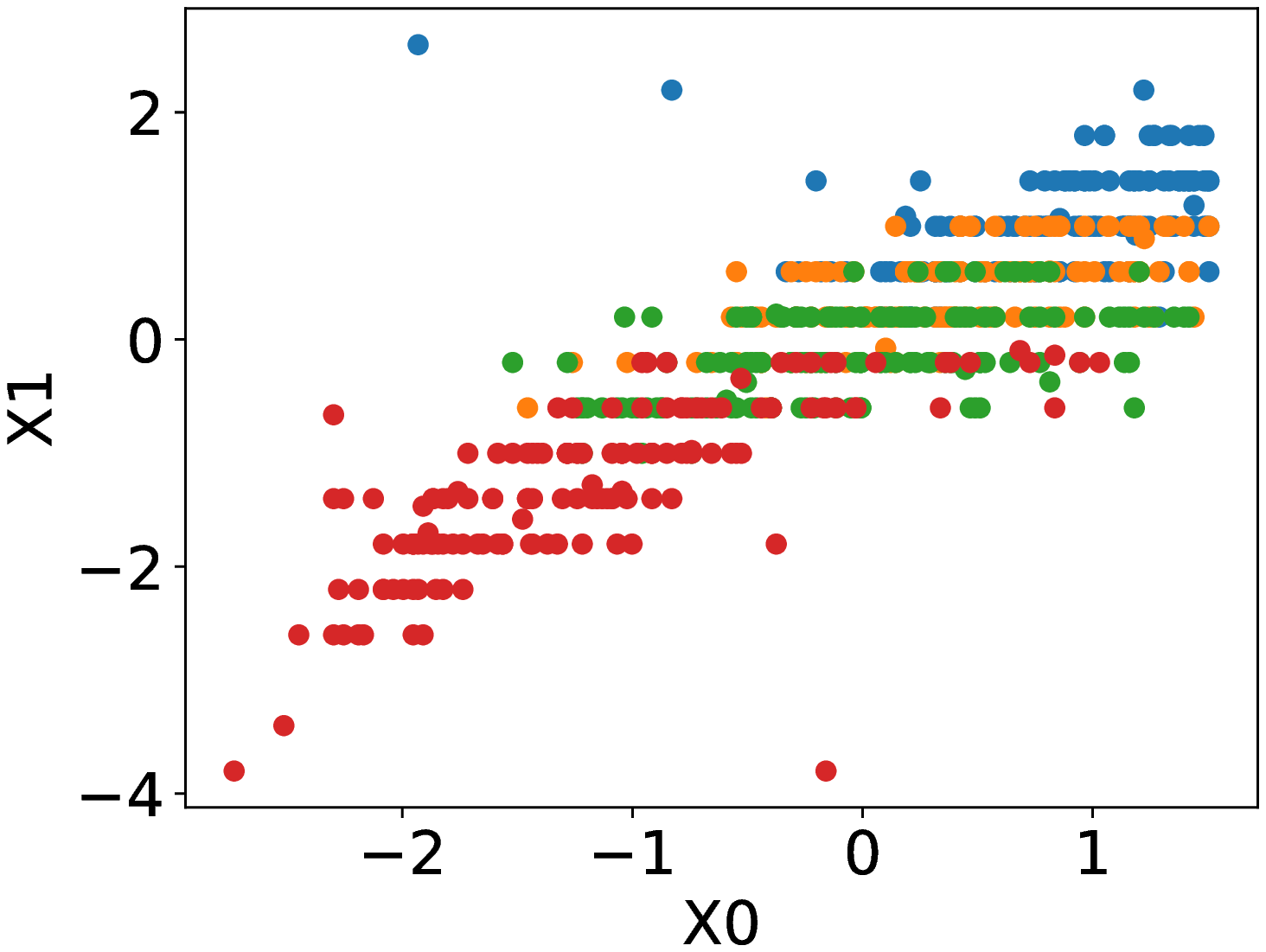}\vspace{-0.1in}
		\end{subfigure}
		\hspace{0.1in}
		\begin{subfigure}[b]{0.22\textwidth}
			\centering
			\includegraphics[width=\textwidth]{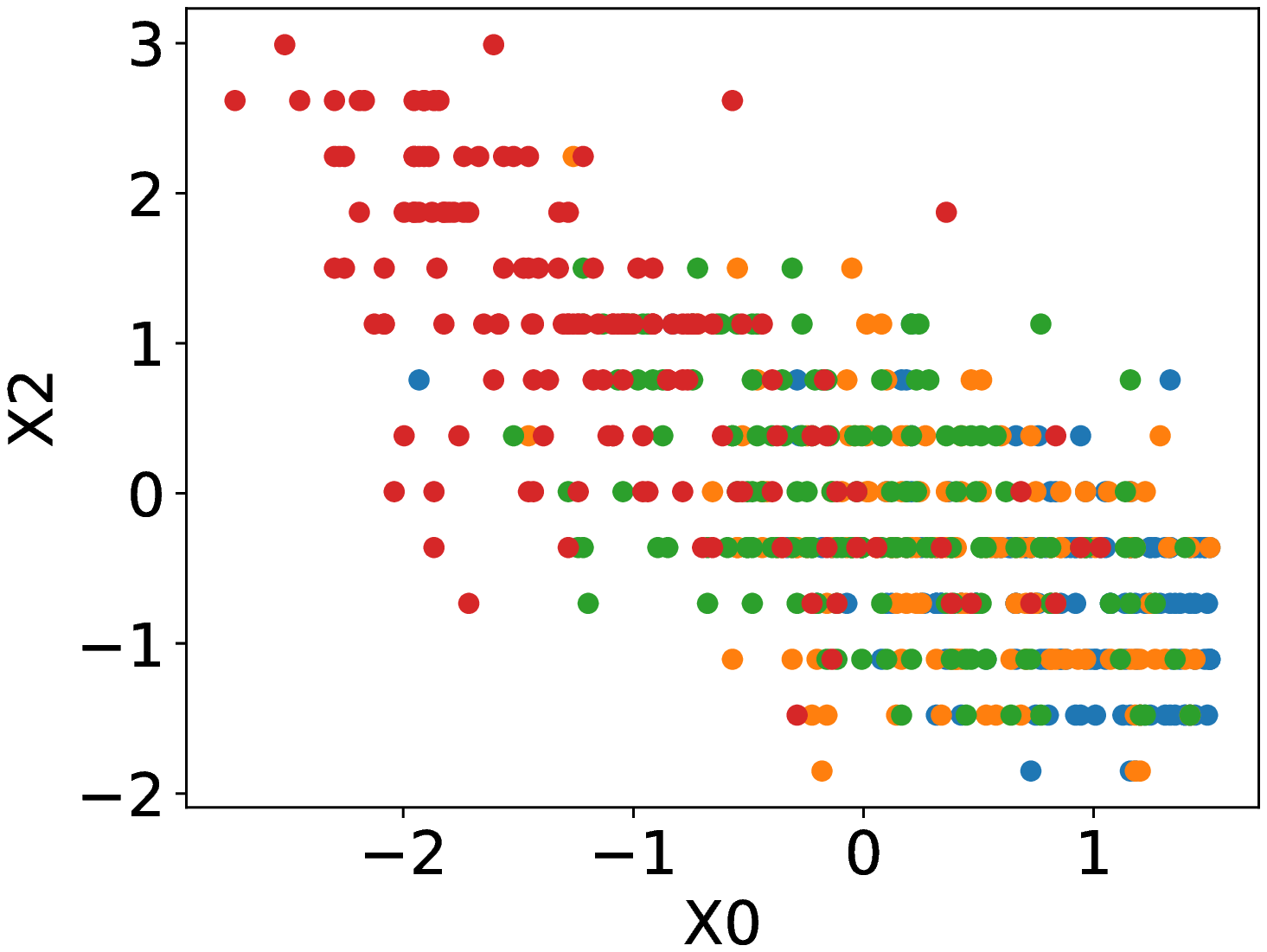}\vspace{-0.1in}
		\end{subfigure}
		\hspace{0.1in}
		\begin{subfigure}[b]{0.22\textwidth}
			\centering
			\includegraphics[width=\textwidth]{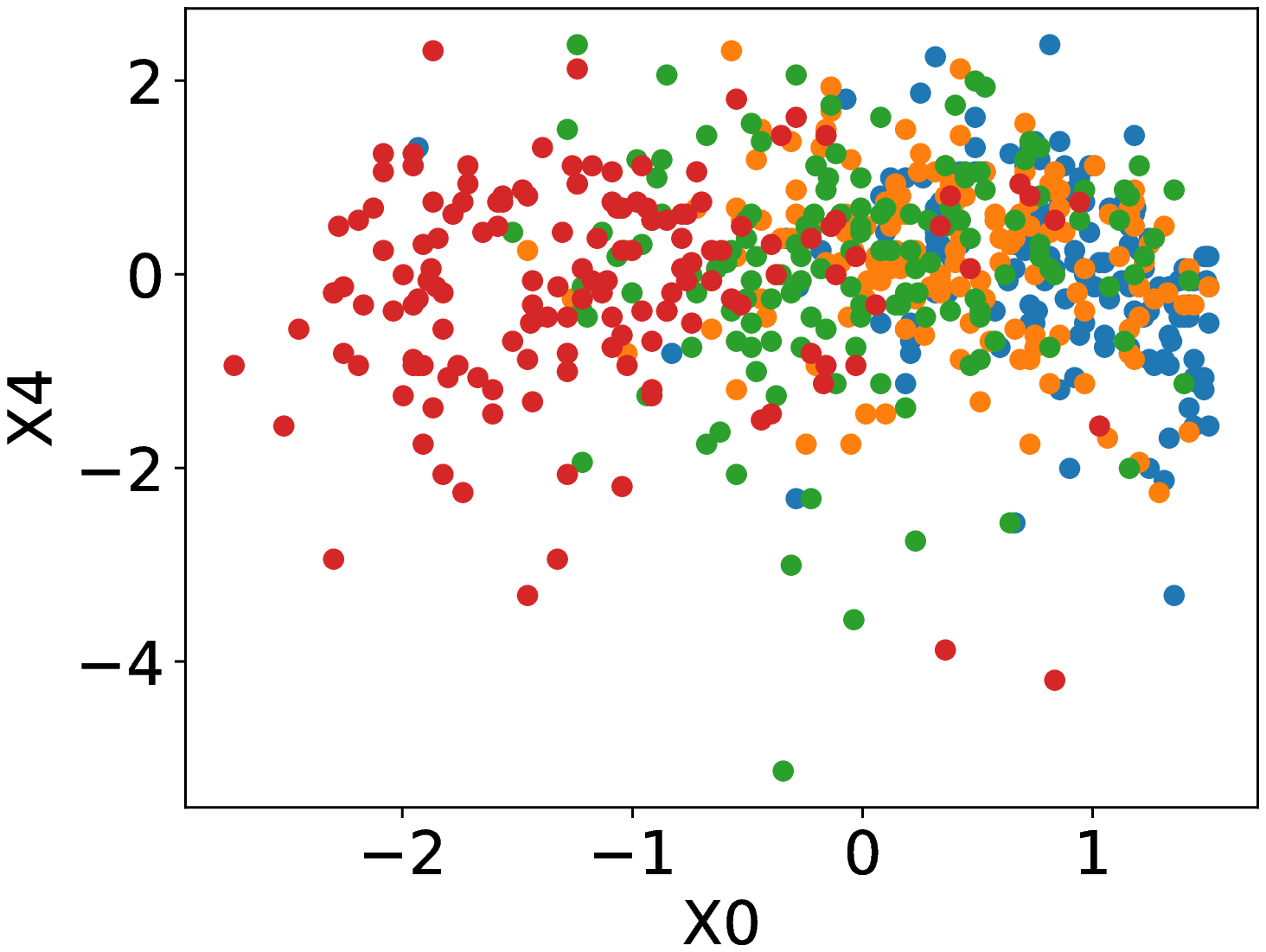}\vspace{-0.1in}
		\end{subfigure}
		\hspace{0.1in}
		\begin{subfigure}[b]{0.22\textwidth}
			\centering
			\includegraphics[width=\textwidth]{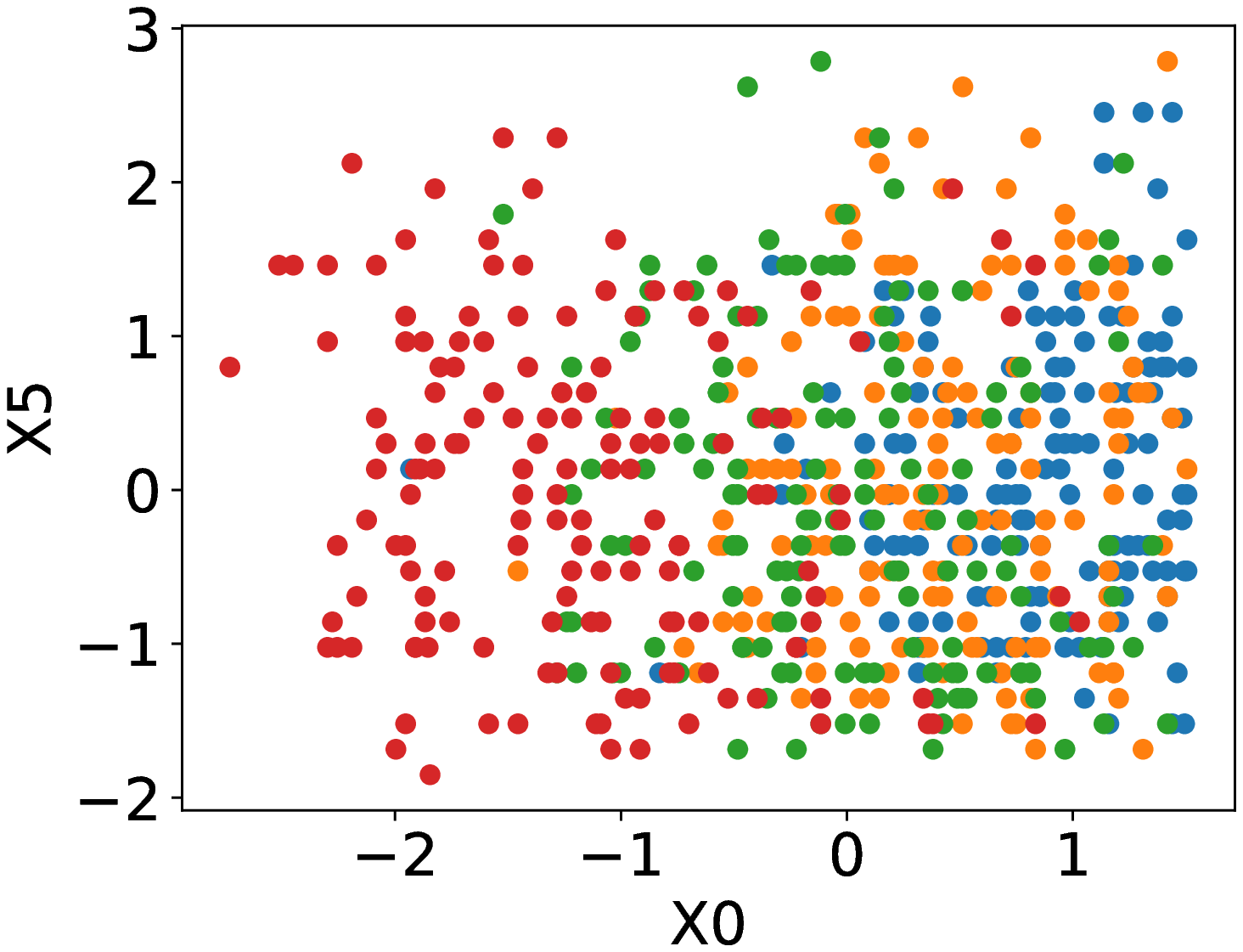}\vspace{-0.1in}
		\end{subfigure}
		\caption{Subgroups in four different intervals of quantiles.}
		\label{fig:subgroups}
	\end{figure}

	One naive way to construct subgroups is dividing the covariate space with respect to the quantiles of estimated ITEs. However, this approach fails to satisfy the essential requirement of subgroup analysis: interpretability. The estimates from a black-box ITE estimator are non-interpretable. Similarly, the subgroups defined based on the estimated quantiles give no explanation (in terms of input covariates) regarding why the samples are assigned to a particular subgroup. To demonstrate this problem clearly, in Fig. \ref{fig:subgroups}, we divide the covariate space of the IHDP dataset into four subgroups based on the intervals of quantiles of CMGP, $[0,25),[25,50),[50,75)$ and $[75,100]$. 
	The colours indicate which subgroup each sample belongs to.
	We can see that the quantile fails to provide interpretability in terms of the input covariates.
	In contrast, R2P constructs easy-to-interpret subgroups based on tree-structure.
	
	\subsection{Complete Results of Table 1 in the Manuscript}
	Below we provide the complete results of Table 1 in the main manuscript.
	We can see that R2P satisfies the target coverage for all datasets even with the narrower confidence interval widths.

	\subsection{Impact of Hyper-Parameters}
	
	Here we show the impact of the hyper-parameters $\gamma$ and $\lambda$ of R2P to the performance.
	We provide the results of varying the hyper-parameters as $\gamma\in\{0.01, 0.02, 0.05, 0.1, 0.15, 0.2, 0.5\}$. and $\lambda\in\{0.1,0.2,0.3,0.4,0.5,0.6,0.7,0.8,0.9\}$.
	We use the same experiment setup in the main manuscript, and repeat the experiments 50 times for each hyper-parameter.

\begin{figure}[ht]
		\vspace{0.2in}
		\centering
		\begin{subfigure}[b]{1.0\textwidth}
			\centering
			\includegraphics[width=\textwidth]{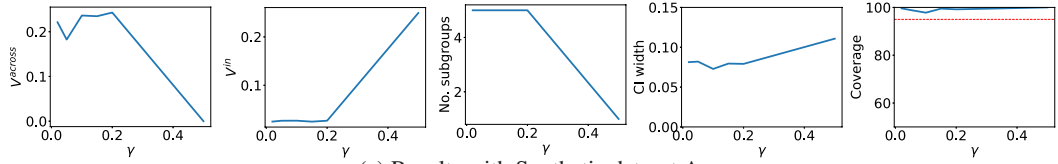}\vspace{-0.1in}
			\caption{Results with Synthetic dataset A.}
			\label{fig:var_gamma_PNAS2}
			\vspace{0.1in}
		\end{subfigure}
		\hfill
		\begin{subfigure}[b]{1.0\textwidth}
			\centering
			\includegraphics[width=\textwidth]{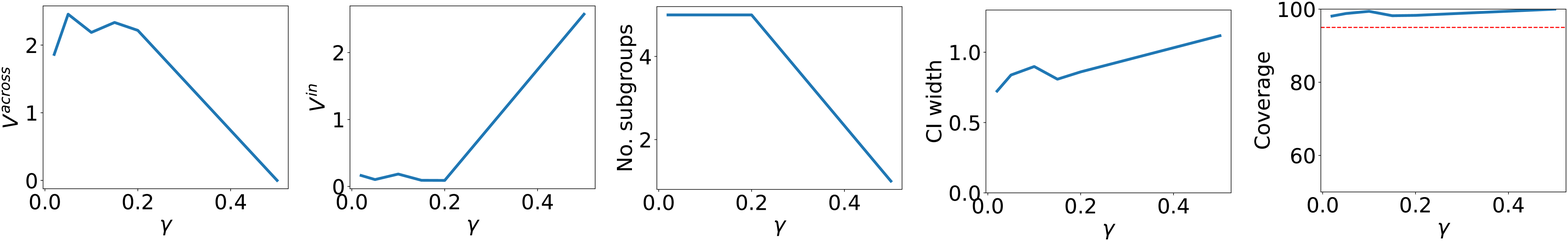}\vspace{-0.1in}
			\caption{Results with Synthetic dataset B.}
			\label{fig:var_gamma_COVID}
			\vspace{0.1in}
		\end{subfigure}
		\hfill
		\begin{subfigure}[b]{1.0\textwidth}
			\centering
			\includegraphics[width=\textwidth]{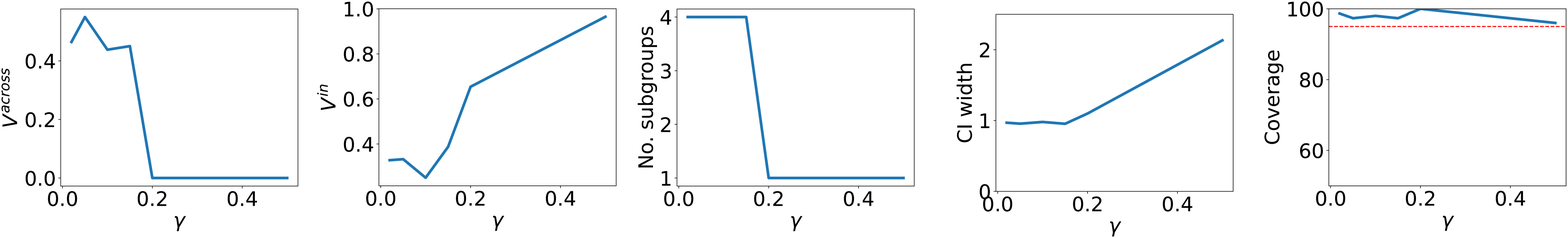}\vspace{-0.1in}
			\caption{Results with IHDP dataset.}
			\label{fig:var_gamma_IHDP}
		\end{subfigure}
		\hfill
		\begin{subfigure}[b]{1.0\textwidth}
			\centering
			\includegraphics[width=\textwidth]{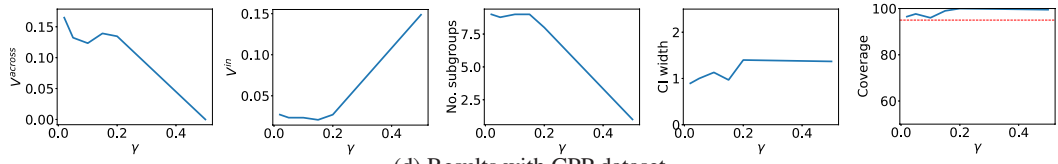}\vspace{-0.1in}
			\caption{Results with CPP dataset.}
			\label{fig:var_gamma_ACIC}
		\end{subfigure}
		\caption{Results on varying $\gamma$. (the red dotted line corresponds to the target coverage rate).
			\label{fig:var_gamma}}
	\end{figure}

	
	\begin{paragraph}{Impact of hyper-parameter $\gamma$}
		We show the impact of the hyper-parameter $\gamma\in[0,1)$ of R2P in Fig. \ref{fig:var_gamma}.
		The hyper-parameter $\gamma$ controls the regularization in R2P.
		From the figures, we can see that as $\gamma$ increases, the number of subgroups converges to one since a group is barely partitioned with the larger $\gamma$.
		This causes the degradation of performance in the following aspects:
		$V^{\textrm{across}}$ decreases, $V^{\textrm{in}}$ increases, and the confidence interval width increases generally.
		Thus, it seems that a smaller value is a better choice for $\gamma$.
		However, if $\gamma$ is too small, the subgroups (i.e., the partition) constructed by R2P becomes overfitted.
		This overfitting issue results in the loss of generalization ability on the unseen data, and a large number of subgroups due to the overfitting makes the subgroup analysis less informative.
		In addition, we can see that our method robustly satisfies the target coverage rate regardless of $\gamma$.
		Therefore, the hyper-parameter $\gamma$ should be appropriately chosen in practice.
	\end{paragraph}
	
	\begin{table}[t]
		\caption{Complete results of Table 1 in the manuscript.}
		\fontsize{8pt}{8pt}\selectfont
		\begin{center}
			\setlength\tabcolsep{2pt}
			\begin{tabular}{c|c|c|c|c|c|c|c|c|c|c}
				\toprule
				& \multicolumn{5}{c|}{Synthetic dataset A} 
				& \multicolumn{5}{c}{Synthetic dataset B} \\
				\midrule
				& $V^{\textrm{across}}$ & $V^{\textrm{in}}$  & \# SGs   & CI width & Cov. (\%)
				& $V^{\textrm{across}}$ & $V^{\textrm{in}}$  & \# SGs   & CI width & Cov. (\%)  \\
				\midrule
				\textsc{R2P}      & \textbf{0.22$\pm$.01} & \textbf{0.03$\pm$.001} & 4.9$\pm$.16 & \textbf{0.08$\pm$.003}  & 98.98$\pm$.24
				& \textbf{2.39$\pm$.04} & \textbf{0.12$\pm$.01} & 5.0$\pm$.16 & \textbf{0.88$\pm$.06} & 98.86$\pm$.23 \\
				\textsc{CCT} 	  & 0.18$\pm$.02 & 0.05$\pm$.01 & 4.4$\pm$.24 & 7.42$\pm$.48  & 100.0$\pm$.00
				& 1.97$\pm$.14 & 0.58$\pm$.15 & 5.0$\pm$.13 & 5.95$\pm$.59  & 99.86$\pm$.13 \\
				\textsc{CT-A}     & 0.19$\pm$.02 & 0.04$\pm$.01 & 4.7$\pm$.21 & 3.96$\pm$.16  & 99.99$\pm$.02
				& 2.24$\pm$.06 & 0.30$\pm$.05 & 5.1$\pm$.15 & 2.77$\pm$.20  & 97.73$\pm$.76 \\
				\textsc{CT-H}     & 0.12$\pm$.03 & 0.11$\pm$.02 & 3.1$\pm$.39 & 4.39$\pm$.22  & 99.98$\pm$.02
				& 2.07$\pm$.13 & 0.53$\pm$.13 & 4.5$\pm$.15 & 3.38$\pm$.32  & 98.20$\pm$.73 \\
				\textsc{CT-L}     & 0.12$\pm$.02 & 0.10$\pm$.02 & 2.9$\pm$.35 & 5.22$\pm$.02  & 99.97$\pm$.06
				& 0.80$\pm$.26 & 1.77$\pm$.27 & 3.1$\pm$.28 & 6.92$\pm$.53  & 99.44$\pm$.47 \\
				\midrule
				\midrule
				& \multicolumn{5}{c|}{IHDP dataset}
				&  \multicolumn{5}{c}{CPP dataset} \\
				\midrule
				& $V^{\textrm{across}}$ & $V^{\textrm{in}}$  & \# SGs   & CI width & Cov. (\%)
				& $V^{\textrm{across}}$ & $V^{\textrm{in}}$  & \# SGs   & CI width & Cov. (\%)  \\
				\midrule
				\textsc{R2P}      & \textbf{0.46$\pm$.04} & \textbf{0.38$\pm$.03} & 4.1$\pm$.12 & \textbf{1.27$\pm$.22} & 97.93$\pm$.39
				& \textbf{0.06$\pm$.02} & \textbf{0.10$\pm$.01} & 5.7$\pm$.30 & \textbf{1.11$\pm$.13} & 98.52$\pm$.34\\
				\textsc{CCT} 	  & 0.30$\pm$.04 & 0.53$\pm$.05 & 4.3$\pm$.13 & 5.70$\pm$.23 & 99.59$\pm$.12
				& 0.03$\pm$.02 & 0.12$\pm$.01 & 6.4$\pm$.20 & 3.60$\pm$.12 & 99.54$\pm$.23 \\
				\textsc{CT-A}     & 0.31$\pm$.04 & 0.57$\pm$.05 & 4.1$\pm$.08 & 3.71$\pm$.08 & 97.41$\pm$.42
				& 0.03$\pm$.01 & 0.12$\pm$.01 & 6.6$\pm$.18 & 2.45$\pm$.06 & 96.60$\pm$.50 \\
				\textsc{CT-H}     & 0.28$\pm$.05 & 0.56$\pm$.05 & 3.8$\pm$.14 & 3.76$\pm$.14 & 97.76$\pm$.40
				& 0.01$\pm$.00 & 0.14$\pm$.01 & 5.2$\pm$.23 & 2.67$\pm$.06 & 98.01$\pm$.40 \\
				\textsc{CT-L}     & 0.27$\pm$.06 & 0.64$\pm$.05 & 2.8$\pm$.23 & 4.75$\pm$.15 & 98.97$\pm$.30
				& 0.01$\pm$.01 & 0.14$\pm$.01 & 2.9$\pm$.29 & 3.23$\pm$.07 & 99.49$\pm$.23 \\
				\bottomrule
			\end{tabular}
		\end{center}
	\end{table}

\begin{figure}[ht]
		\vspace{0.2in}
		\centering
		\begin{subfigure}[b]{1.0\textwidth}
			\centering
			\includegraphics[width=\textwidth]{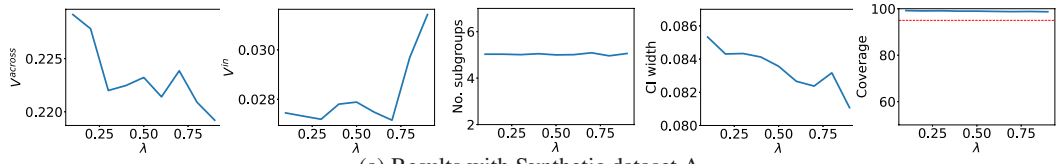}\vspace{-0.1in}
			\caption{Results with Synthetic dataset A.}
			\label{fig:var_lambda_PNAS2}
			\vspace{0.1in}
		\end{subfigure}
		\hfill
		\begin{subfigure}[b]{1.0\textwidth}
			\centering
			\includegraphics[width=\textwidth]{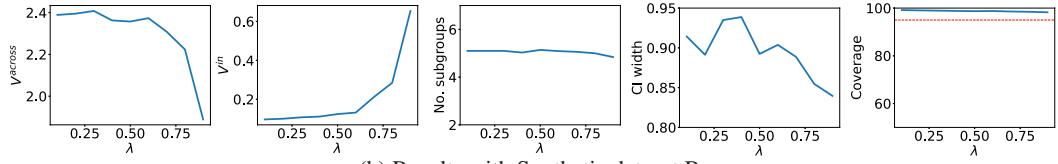}\vspace{-0.1in}
			\caption{Results with Synthetic dataset B.}
			\label{fig:var_lambda_COVID}
			\vspace{0.1in}
		\end{subfigure}
		\hfill
		\begin{subfigure}[b]{1.0\textwidth}
			\centering
			\includegraphics[width=\textwidth]{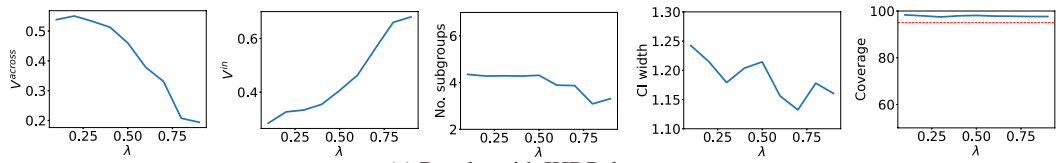}\vspace{-0.1in}
			\caption{Results with IHDP dataset.}
			\label{fig:var_lambda_IHDP}
		\end{subfigure}
		\caption{Results on varying $\lambda$. (the red dotted line corresponds to the target coverage rate).
			\label{fig:var_lambda}}
	\end{figure}

	\begin{paragraph}{Impact of hyper-parameter $\lambda$}
		We show the impact of the hyper-parameter $\lambda$ of R2P in Fig. \ref{fig:var_lambda}.
		The hyper-parameter $\lambda\in[0,1]$ controls the weight between the homogeneity within each subgroup and the confidence interval discrimination in the confident criterion.
		With smaller $\lambda$, the homogeneity within each subgroup is more emphasized in the criterion.
		So in that case, from the figures, we can see that R2P finds the larger number of subgroups, which results in the higher $V^{\textrm{across}}$ and the lower $V^{\textrm{in}}$.
		On the other hand, with larger $\lambda$, the confidence interval discrimination is weighted higher in the criterion, and thus, the confidence interval width decreases generally.
		We can see that our method robustly satisfies the target coverage rate regardless of $\lambda$ in most cases.
		Overall, $\lambda$ should be appropriately chosen considering the variance over the entire population.
	\end{paragraph}
	
\end{document}